\documentclass{article}

\PassOptionsToPackage{numbers, sort&compress}{natbib}

\usepackage[main, final]{neurips_2026}

\usepackage[utf8]{inputenc}
\usepackage[T1]{fontenc}
\usepackage{hyperref}
\usepackage{url}
\usepackage{booktabs}
\usepackage{amsfonts}
\usepackage{nicefrac}
\usepackage{microtype}
\usepackage[dvipsnames]{xcolor}
\usepackage{paralist}

\usepackage{amsmath,amssymb}
\usepackage{amsthm}
\usepackage{cleveref}
\usepackage{graphicx}
\usepackage{graphics}
\graphicspath{{../}}
\usepackage{wrapfig}
\usepackage{caption}
\usepackage{subcaption}
\usepackage{multicol}
\usepackage{tabularx}
\usepackage{diagbox}
\usepackage{siunitx}
\usepackage{algorithm}
\usepackage{algorithmic}
\usepackage{xspace}
\usepackage{cancel}
\usepackage{verbatim}
\usepackage{multirow}

\usepackage{tikz,ifthen}
\usetikzlibrary{arrows,trees,backgrounds,automata,shapes,decorations,plotmarks,fit,calc,positioning,shadows,chains}

\tikzstyle{every pin edge}=[<-,shorten <=1pt]
\tikzstyle{neuron}=[circle,fill=black!25,minimum size=17pt,inner sep=0pt]
\tikzstyle{input neuron}=[neuron, fill=green!40]
\tikzstyle{output neuron}=[neuron, fill=red!40]
\tikzstyle{hidden neuron}=[neuron, fill=blue!40]
\tikzstyle{constructed neuron}=[neuron, fill=orange!50]
\tikzstyle{annot} = [text width=6em, text centered]
\tikzstyle{nnedge} = [-{stealth},shorten >=0.1cm, shorten <=0.05cm,line width=0.8pt,black]

\newtheorem*{theorem*}{Theorem}
\newtheorem{theorem}{Theorem}

\theoremstyle{definition}
\newtheorem{definition}{Definition}[section]

\crefname{algorithm}{Alg.}{Algs.}
\crefname{figure}{Fig.}{Figs.}
\crefname{section}{Sec.}{Secs.}
\crefname{theorem}{Thm.}{Thms.}
\crefname{table}{Tab.}{Tabs.}
\crefname{appendix}{App.}{Apps.}

\hypersetup{
	hidelinks,
	breaklinks=true,
	colorlinks=true,
	urlcolor=blue,
	bookmarksopen=false,
}

\newcommand{\sat}{\texttt{UNSAFE}}
\newcommand{\unsat}{\texttt{SAFE}}
\newcommand{\unsafe}{\texttt{UNSAFE}}
\newcommand{\safe}{\texttt{SAFE}}
\newcommand{\unknown}{\texttt{UNKNOWN}}

\newcommand{\nn}{\mathcal{N}}

\newcommand{\pave}{\textit{PAVE}\xspace{}}
\newcommand{\RETURN}{\STATE \textbf{return} }

\usepackage{amsmath,amsfonts,bm}

\def\eqref#1{equation~\ref{#1}}

\def\1{\bm{1}}

\DeclareMathAlphabet{\mathsfit}{\encodingdefault}{\sfdefault}{m}{sl}
\SetMathAlphabet{\mathsfit}{bold}{\encodingdefault}{\sfdefault}{bx}{n}

\title{Pareto DNN Verification: Fast for Most Queries}

\author{%
	\makebox[0.45\textwidth][c]{\begin{tabular}[t]{c}
	Yizhak Y. Elboher\thanks{Equal contribution} 
	\\
	\normalfont
	The Hebrew University of Jerusalem\\ \normalfont Masaryk University
	\\
	\texttt{yizhak.elboher@mail.huji.ac.il}
	\end{tabular}}
	\And
	\makebox[0.45\textwidth][c]{\begin{tabular}[t]{c}
	Avraham Raviv\footnotemark[1]
	\\
	\normalfont
	Bar Ilan University
	\\
	\texttt{avraham.raviv@biu.ac.il}
	\end{tabular}}
	\And
	\makebox[0.45\textwidth][c]{\begin{tabular}[t]{c}
	Amihay Elboher
	\\
	\normalfont
	Ben Gurion University
	\\
	\texttt{amihay@bgu.ac.il}
	\end{tabular}}
	\And
	\makebox[0.45\textwidth][c]{\begin{tabular}[t]{c}
	Zhouxing Shi
	\\
	\normalfont
	University of California, Riverside\\
	\texttt{zhouxing.shi@ucr.edu}
	\end{tabular}}
	\And
	\makebox[0.45\textwidth][c]{\begin{tabular}[t]{c}
	Hillel Kugler
	\\
	\normalfont
	Bar Ilan University
	\\
	\texttt{hillelk@biu.ac.il}
	\end{tabular}}
	\And
	\makebox[0.45\textwidth][c]{\begin{tabular}[t]{c}
	Guy Katz
	\\
	\normalfont
	The Hebrew University of Jerusalem\\
	\texttt{g.katz@mail.huji.ac.il}
	\end{tabular}}
}

\begin{document}

	\maketitle
	
	
	\begin{abstract}
		\label{sec:abstract}
		Formal verification of deep neural networks (DNNs) faces a fundamental 
		scalability challenge: verifying robustness is NP-hard and 
		state-of-the-art tools often time out on modern architectures. 
		We propose \emph{PAreto VErification}, or \pave{}: by augmenting 
		DNNs with early exits (EEs) and verifying exit-by-exit, we solve a 
		large fraction of robustness queries---roughly 80\% in our 
		evaluation---in a small fraction of the time required for direct 
		verification. The key insight is that many inputs in a robustness 
		neighborhood exit at an early layer. Verifying this early-exit behavior is often a tractable problem,
		 allowing verification to terminate long before the full network is 
		 analyzed. EEs can be added without modifying the base network, 
		 and incur at most a 0.3\% accuracy change across our tested 
		 architectures. We formalize the novel robustness property for EE 
		 networks, prove it is Fixed-Parameter Tractable under a trace 
		 stability condition (which commonly holds in practice), and present 
		 a sound and complete verification algorithm. Two sound 
		 optimizations---\emph{break} (early termination when the winner 
		 dominates) and \emph{continue} (skipping the inner loop when no 
		 runner-up can win)---reduce SAFE-case verification time by up to 
		 $10\times$. Experiments on MNIST, CIFAR-10, and CIFAR-100 with 
		 networks from 600K to 33M parameters demonstrate that \pave{} 
		 enables solving queries that were not solved otherwise, including 
		 formal verification of ResNet-18 on CIFAR-100.
	\end{abstract}

	\section{Introduction}\label{sec:intro}
	
	\paragraph{The Scalability Crisis in DNN Verification.}
	Deep Neural Networks (DNNs) are increasingly deployed in safety-critical domains---from medical diagnostics~\citep{HUPaJeLuYeCh23} to autonomous driving---where formal guarantees of correct behavior are essential. Formal verification~\citep{KaBaDiJuKo17,Eh17,TjXiTd20,WaZhXuLiJaHsKo21} provides such guarantees by checking whether a network's output satisfies a specified property for \emph{all} inputs in a given domain, thereby ruling out adversarial perturbations~\citep{SzZaSuBrErGoFe14} and safety violations~\citep{DaSkBeRuTeSeOmSzGoAm24,Ru22}. However, the problem is NP-hard in general~\citep{KaBaDiJuKo17}, and even the most advanced tools~\citep{WaZhXuLiJaHsKo21,XuZhWaWaJaLiHs21} struggle to scale beyond small networks, often timing out on architectures like ResNet-18 or VGG-16. This scalability bottleneck is one of the most pressing open problems in the field~\citep{LiArLaStBaKo21,BrMuBaJoLi23}.
	
	\paragraph{The Pareto Verification Insight.}
	A well-known property of DNN inference is that many inputs---particularly easy, well-separated ones---can be classified confidently after only a few layers of computation. Early exit (EE) architectures~\citep{TeMcKu16,WaLuCrTuYuGo17} formalize this intuition by inserting intermediate classifiers at various depths; an input exits early whenever its prediction confidence at an intermediate layer exceeds a threshold, saving computation without sacrificing accuracy. We ask: \emph{can this same depth-shortcutting insight accelerate formal verification?}
	
	We answer this question affirmatively, through the \pave{} framework: by augmenting DNNs with early exits and verifying exit-by-exit, we resolve a large fraction of robustness queries---roughly 80\% in our evaluation---in a fraction of the time, since many inputs in a robustness neighborhood are handled at an early exit. For a given input~$x$ and perturbation radius~$\epsilon$, if all perturbed inputs $x'\in B_\epsilon^x$ exit at the same layer as~$x$ (a condition we call \emph{trace stability}), then verifying robustness reduces to checking only the sub-network up to that exit---a strictly smaller and easier problem. We prove that this reduced problem is Fixed-Parameter Tractable (\cref{the:Pee_is_FPT}), and we empirically confirm that trace stability holds in the vast majority of \safe{} cases (\cref{fig:inf-vs-ver-exit-heatmap}), which are commonly the harder ones.
	
	\paragraph{Adding Early Exits without Hurting Accuracy.}
	\pave{} requires the target DNN to have EEs. 
	The key concern is whether adding EEs degrades the network's predictive accuracy 
	or robustness. We show it does not~\cite{RaSrChPaCo24}: 	
	EEs are added as thin, frozen auxiliary classifiers that do not alter the base 
	network's parameters. Trained sequentially on residual examples, they preserve 
	the original model's accuracy to within 0.3\% across all tested architectures 
	(Table~\ref{tab:model_metrics}). Thus, standard DNNs 
	can be augmented with EEs and 
	subjected to pareto versification as a transparent acceleration technique, 
	applicable alongside base or certified training.
	
	\paragraph{A Novel Formal Framework for EE Verification.}
	While the \pave{} paradigm is conceptually intuitive,
	realizing it requires solving a fundamentally new formal problem. Existing verifiers assume a single fixed output layer; they cannot handle EE networks, whose execution follows conditional branching semantics: the output is produced at the \emph{first} exit where the confidence threshold is met. A naive encoding may produce spurious counterexamples where a runner-up wins at exit $e$ while the true winner already prevailed at an earlier exit $e' < e$.
	
	We address this by formalizing a novel robustness property $P_{ee}$ (\cref{subsec:3.1}) that explicitly conditions each exit on the absence of earlier winners. We present a baseline sound and complete verification algorithm (\cref{alg:basic-early-exit-verification}), then observe that it may issue many redundant queries when the answer is \safe{}. We introduce two independent sound optimizations that together yield up to $10\times$ speedup on \safe{} cases (\cref{alg:combined-break-then-continue-optimizations}):
	\begin{inparaenum}[(i)]
		\item \textbf{Break}: if the winner always exceeds the threshold at exit $k$ and no prior counterexample exists, verification terminates immediately;
		\item \textbf{Continue}: if the winner always exceeds $1-T$, no runner-up can win at exit $k$, so the inner per-class loop is skipped.
	\end{inparaenum}
	Both optimizations maintain soundness and completeness (proofs appear in \cref{appendix:Proofs}).
	
	\paragraph{Contributions.}
	In this work, we contribute to the formal verification of DNNs by:
	\begin{inparaenum}[(i)]
		\item introducing \pave{} as a general paradigm for accelerating DNN verification---solving a large fraction of queries in a fraction of the time---and showing how EEs enable it without changing the backbone;
		\item formalizing the EE robustness property $P_{ee}$ and proving it is Fixed-Parameter Tractable under trace stability;
		\item presenting a sound and complete baseline algorithm and two sound, efficiency-improving optimizations;
		\item conducting extensive experiments on multiple benchmarks---MNIST, CIFAR-10, CIFAR-100 across FC, CNN, ResNet-18, and VGG-16 architectures---including queries for which the baseline times out; and
		\item analyzing the interplay between exit threshold, exit placement, accuracy, and verifiability to give practitioners design guidance.
	\end{inparaenum}

	\section{Preliminaries}\label{sec:prelim}
	
	\subsection{Notations}
	A DNN is represented as a function $\nn: \mathbb{R}^n \to \mathbb{R}^m$, where $n,m\in\mathbb{N}$ are input and output dimensions, respectively.
	For an input $\mathbf{x} \in \mathbb{R}^n$, $\nn(\mathbf{x})$ outputs a vector $\mathbf{y} \in \mathbb{R}^m$. We focus on classification networks, where the predicted class is the index of the highest value in $\mathbf{y}$ (the \textit{winner}); other indices are \textit{runner-ups}.
	An $\epsilon$-ball around $x$, denoted $B_\epsilon^x$, is the set $\{ x' \in \mathbb{R}^n : \|x' - x\|_p \leq \epsilon \}$, 
    where \(\Vert{}\cdot\Vert{}_p\) is the \(\ell _{p}\) norm defined by a given parameter $p\in [1,\infty)$. For $k\in\mathbb{N}$, we write $[k]=\{1,\dots,k\}$.
	
	\subsection{Formal Verification of DNNs}
	The formal verification of a DNN \( \nn : \mathbb{R}^n \to \mathbb{R}^m \) can be cast into a constraint satisfiability problem,
	where the goal is to determine whether a property $P$ is \textit{satisfiable} in $\nn$. $P$ represents the existence of an input to $\nn$ within a specific domain $\mathcal{D}$ whose output satisfies a constraint \( \phi \):
	\[
	\exists \mathbf{x} \in \mathcal{D} \text{ such that } \phi(\mathbf{x}, \nn(\mathbf{x})) \text{ is satisfied}.
	\]
	Typically, \( \phi \) captures undesirable behavior by encoding the negation of a desired property. Thus, if $P$ is satisfiable, $\nn$ is said to be \unsafe{}.
If $P$ is not satisfiable, the desired property holds and $\nn$ is \safe{}.
	
	\subsection{DNNs with Early Exits}
	A DNN with early exits is a network augmented with additional decision points, known as \textit{exits}, within its architecture. These exits allow the network to terminate inference early if a condition, typically a confidence threshold, is met, reducing computational cost while maintaining accuracy.
	Let \( \nn_{ee}: \mathbb{R}^n \to \mathbb{R}^m \) denote a network with \( k \) exits, and let \( \mathbf{y}^{(j)} \) represent the neuron values of the \( j \)-th exit. Inference at exit \( j \) terminates if $f(\mathbf{y}^{(j)}) \geq T_j$,
	where \( T_j \) is a predetermined confidence threshold for the \( j \)-th exit, and \( f \) is a confidence function.
	In early works~\citep{TeMcKu16}, \( f(\mathbf{y}^{(j)}) \) was computed as the entropy of the $j$'th exit logits; later, alternative gating mechanisms were proposed, including maximum SoftMax probability~\citep{PaSeRo16}, confidence accumulation~\citep{ScScBaUn20}, and learned exit conditions~\citep{XiTaYuLi21}.
	Regardless of the gating function \( f \), the final output \( \mathbf{y} \) is determined by the first exit satisfying the threshold; if no threshold is met, the output comes from the last exit.
	In this work, we use a fully connected layer followed by a \textit{SoftMax} activation as the confidence function \(f\), with the condition \( \max(\mathbf{y}^{(j)}) > T_j \).
	A common threshold, also used in many of our experiments, is \( T = 0.9 \)~\citep{RaSrChPaCo24,RoYuJu22}. \cref{fig:fullyConnectedNetworkWithExitsAbove} in \cref{app:early-exit-example-figure} depicts a DNN with EEs.

	\section{The \pave{} Framework}
	\label{sec:method}
	
	\pave{} decomposes the hard problem of verifying a deep network into a sequence of smaller sub-problems, one per exit, exploiting the fact that for typical inputs the verification terminates early. This section formalizes the approach. We begin in \cref{subsec:3.1} by redefining local robustness to accommodate multiple exits. Then, in \cref{subsec:3.2}, we present the base algorithm that mirrors EE conditional inference semantics. In \cref{subsec:3.3}, we analyze the complexity of this approach and show that, under trace stability, its cost is significantly reduced. Finally, in \cref{subsec:3.4}, we introduce two optimizations that reduce redundant queries while preserving soundness and completeness.
	
	\paragraph{Why Standard Verifiers Cannot Handle EE Networks.}
	Networks with EEs introduce two challenges that break standard verification:
	\begin{inparaenum}[(1)]
		\item In conventional DNNs, the output property is defined on a single output layer; for EE networks, this must be adapted for multiple output layers.
		\item Multiple exits introduce conditional branching: if an exit is confident it returns; otherwise computation continues. Naive verification ignores this, producing spurious counterexamples where a runner-up wins at exit $e$ while the winner already won at $e' < e$.
	\end{inparaenum}
	\pave{} resolves both issues by formalizing the robustness query to respect EE semantics and leveraging EEs to shorten verification.
	
	\subsection{Revised Robustness Property} \label{subsec:3.1}
	For a standard DNN $\nn$ (without EEs), the property to negate local robustness around a sample $x$ is:
	\begin{equation*}
			\begin{aligned}
				P&:=\exists x'\in B^x_\epsilon, \exists i \in \mathcal{C} \text{ such that } \nn(x')_i > \nn(x')_w
			\end{aligned}
	\end{equation*}
	where $\mathcal{C}$ is the set of possible output labels $\{1,\dots,m\}$, $B^x_\epsilon$ is the $\epsilon$-ball around $x$, $w$ is the index of the winner class and $\nn(x)_j$ is the $j$'th value in $\nn(x)$.
	
	In DNNs with EEs, the inference process enables outputs to be returned from various exits, corresponding to different neurons. This adds ambiguity to the traditional definition: the verification process must condition the validity of the specification on the assumption that a specific exit serves as the actual output layer for the given input.
	A counterexample to robustness of a network with EEs is one where (i) a ``runner-up output'' wins at an early or output exit \( e \), and (ii) the desired output does not prevail at any exit preceding \( e \).
	These conditions are encapsulated in the following property \( P_{ee} \), which defines the negation of the revised local robustness property for a DNN with EEs. The indices of the layers with EEs and the index of the output layer are denoted with $ee$ and $last$, respectively.
	\begin{equation*}
			\begin{aligned}
				P_{ee} := &\ \exists x'\in B^x_\epsilon, i\in\mathcal{C}\setminus \{w\},e\in ee\cup \{last\}:\\
				&\ \big((\nn(x')^e_i > T_e \land e<last) \ \lor\  (\nn(x')_i > \nn(x')_w \land e=last)\big) \ \ \land \\
				&\ \forall j\in ee\cap\{1,\dots, e-1\}: \nn(x')^j_w < T_j
			\end{aligned}
	\end{equation*}
	Here, \( P_{ee} \) asserts that (first line) there exists an input $x'$, a runner-up $i$, and an exit $e$ such that (second line) the runner-up wins at the early exit (left side) or at the last layer (right side), and (third line) the winner does not prevail at any earlier exit. If \( P_{ee} \) is satisfiable, the network is \unsafe{} for robustness within $B_\epsilon^x$. Otherwise, $\nn$ is \safe{}.

	\subsection{Verification Algorithm} \label{subsec:3.2}
	To verify robustness in DNNs with EEs, we propose \cref{alg:basic-early-exit-verification}.
	The algorithm iterates through all exits (outer loop). For each exit, it examines each runner-up class (inner loop) to determine whether there exists a counterexample where the runner-up wins and the output is produced at the current exit. This is accomplished by a verification query (line 5 or 7) that tries to satisfy: the runner-up wins, and the winner has not already won in any preceding exit. If $\sat$ is returned, the resulting example satisfies \( P_{ee} \), providing a counterexample to robustness. Otherwise, if $\unsat$ is returned for all exits and classes, local robustness is verified. The call to Verify on line 9 launches an underlying verification tool to solve a standard query.
	
	\begin{figure}[t]
		\centering
		\begin{minipage}[t]{0.98\textwidth}
			\centering
			\scalebox{0.8}{%
				\begin{minipage}{1.23\linewidth}
					\refstepcounter{algorithm}\label{alg:basic-early-exit-verification}%
					\hrule height 0.8pt
					\vspace{2pt}
					\noindent\textbf{Algorithm \thealgorithm}\quad Pareto Verification of DNN Robustness
					\vspace{2pt}
					\hrule
					\vspace{4pt}
					\noindent\textbf{Input} $\nn$, $x$, $\epsilon$ \quad
					\textbf{Output} $\nn$ is robust in $B^x_{\epsilon}$, counterexample
					\begin{algorithmic}[1]
						\STATE {$w, ee, last$ = $argmax(\nn(x))$, indices of layers with EEs, last layer index}
						\FOR {$k \in ee\cup \{last\}$}
						\FOR{$i \in \mathcal{C}\setminus\{w\}$}
						\IF{k $\neq$ last}
						\STATE {$\mathcal{P} := \exists x'\in B_{\epsilon}^x: (\nn(x')^k_i>T_k) \land (\forall e\in ee \cap [k-1] : \nn(x')^e_w<T_e)$}
						\ELSE
						\STATE{$\mathcal{P} := \exists x'\in B_{\epsilon}^x:(\nn(x')_w < \nn(x')_i) \land (\forall e\in ee: \nn(x')^e_w<T_e)$}
						\ENDIF
						\STATE{res, cex = Verify($\nn, B^{x}_{\epsilon}, \mathcal{P})$}
						\IF{res == \sat}
						\RETURN \sat, cex
						\ENDIF
						\ENDFOR
						\ENDFOR
						\RETURN \unsat
					\end{algorithmic}
					\vspace{4pt}
					\hrule height 0.8pt
				\end{minipage}%
			}%
		\end{minipage}%
		
		\vspace{0.8em}
		\begin{minipage}[t]{0.98\textwidth}
			\centering
			\scalebox{0.8}{%
				\begin{minipage}{1.23\linewidth}
					\refstepcounter{algorithm}\label{alg:combined-break-then-continue-optimizations}%
					\hrule height 0.8pt
					\vspace{2pt}
					\noindent\textbf{Algorithm \thealgorithm}\quad Pareto Verification with Break and Continue Optimizations
					\vspace{2pt}
					\hrule
					\vspace{4pt}
					\noindent\textbf{Input} $\nn$, $x$, $\epsilon_p$ \quad
					\textbf{Output} $\nn$ is robust in $B_{\epsilon}^x$, counterexample
					\begin{algorithmic}[1]
						\STATE {$w, ee, last$ = $argmax(\nn(x))$, indices of layers with EEs, last layer index}
						\FOR {$k \in ee\cup \{last\}$}
						\IF{\textcolor{blue}{k $\neq$ last}}
						\STATE{\textcolor{blue}{$\mathcal{P}:=\exists x'\in B_{\epsilon}^x:\nn(x')^k_w<T$}}
						\ELSE
						\STATE{\textcolor{blue}{$\mathcal{P}:=\exists
								x'\in B_{\epsilon}^x,\exists i\in\mathcal{C}\setminus\{w\}: \nn(x')_w<\nn(x')_i
								$}}
						\ENDIF
						\STATE {\textcolor{blue}{res, cex = Verify($\nn, B^x_\epsilon,\mathcal{P}$)}}
						\IF{\textcolor{blue}{res == $\unsat$}}
						\RETURN {\textcolor{blue}{$\unsat$}}
						\ENDIF
						\STATE {\textcolor{orange}{$res$, $cex$ = Verify($\nn,B^x_\epsilon,\forall x'\in B^x_\epsilon: \nn^k_w(x')<1-T$)}}
						\IF {\textcolor{orange}{k == last $\lor$ res == $\sat$}}
						\STATE {res, cex = $ExistsPrevCEX(\nn,x,\epsilon,k)$}
						\IF {res == $\sat$}
						\RETURN {$\sat$, cex}
						\ENDIF
						\ENDIF
						\ENDFOR
						\RETURN $\unsat$
					\end{algorithmic}
					\vspace{4pt}
					\hrule height 0.8pt
				\end{minipage}%
			}%
		\end{minipage}
	\end{figure}

	\begin{theorem}\label{the:basic_verifies_Pee}
		If the underlying verifier is sound and complete, \cref{alg:basic-early-exit-verification} is sound and complete.
	\end{theorem}

	\pave{} returns \safe{} or \unsafe{} only when all required subqueries are proved; if any required subquery is \unknown{}, \pave{} returns \unknown{}. Therefore, \pave{} remains sound with an underlying incomplete verifier, but may be incomplete itself.
    This limits applicability to operations supported by the chosen verifier: for example, transcendental activations can be used only when the verifier provides a sound encoding or bound for them.
	
	\subsection{Fixed-Parameter Tractable Complexity} \label{subsec:3.3}
	
	\cref{alg:basic-early-exit-verification} contains two nested loops; while it returns immediately upon finding a counterexample (\sat), it must exhaust the entire loop before concluding \unsat. This motivates further improvements, as we show that in some cases, local robustness in DNNs with early exits can be determined more efficiently.
	For that purpose, we define the \textit{trace} of an input and its stability as follows.
	\begin{definition}
		The trace $\tau(x)$ of an input $x$ in DNN $\nn$ with EEs is the set of layers $x$ is propagated through. Given an $\epsilon>0$, $\tau(x)$ is \textit{stable} in $B_\epsilon^x$ if $\forall x'\in B_\epsilon^x: \tau(x')=\tau(x)$.
	\end{definition}
	
	The trace of an input determines which parts of the network must be considered to verify robustness for that input. If all $x' \in B_\epsilon^x$ share the same trace as $x$, then if \cref{alg:basic-early-exit-verification} does not return \sat{} before reaching the exit layer of $x$, it will eventually return \unsat{}. Hence, under trace stability, the complexity of solving $P_{ee}$ depends on $|\tau(x)|$---the number of layers in $\tau(x)$---rather than the total number of layers in $\nn$.
	In \cref{sec:results} (\cref{fig:inf-vs-ver-exit-heatmap}), we show that trace stability often holds in practice.
	
	\begin{definition}[{\citep[Def.~1]{Do12}}]
		A problem is \textit{Fixed-Parameter Tractable} (\textit{FPT}) with respect to a parameter $p$, (denoted \textit{FPT}$(p)$), if it can be solved in time $f(p)\cdot poly(n)$, where $f$ is a computable function of $p$, and $n$ is the input size.
	\end{definition}
	
	\begin{theorem}\label{the:Pee_is_FPT}
		Given a ReLU network $\nn$ with EEs, layer width bound $k$, input $x$, 
		and $\epsilon > 0$, if $\tau(x)$ is \textit{stable} in $B_\epsilon^x$, 
		then solving $P_{ee}$ with $(\nn,x,\epsilon)$ using an exact verifier 
		is \textit{FPT}$(k \cdot |\tau(x)|)$.
	\end{theorem}
	
	This result is the theoretical foundation for \pave{}: when trace 
	stability holds, verifying a robustness query scales with the depth of 
	the \emph{exit prefix} rather than the full network. 
	With an incomplete verifier, the same prefix reduction still applies 
	operationally, but the FPT statement is a bound on the exact decision 
	problem rather than a completeness guarantee for that verifier. 
	Under typical EE training~\citep{TeMcKu16}, the trace of most inputs is 
	significantly shorter than the full network, yielding a substantial 
	reduction in worst-case complexity.

	\subsection{Two Sound Optimizations} \label{subsec:3.4}
	Proving \unsat{} with \cref{alg:basic-early-exit-verification} requires calling Verify for every exit and every class. We discuss two improvements to expedite the overall procedure. To maintain readability, we denote lines 3--13 in the algorithm as $ExistsPrevCEX(\nn, x, \epsilon, k)$, which rules out adversarial examples in earlier layers.
	
	\paragraph{Continue Optimization (\textcolor{orange}{orange} lines in \cref{alg:combined-break-then-continue-optimizations}).}
	One improvement reduces the number of queries per exit by checking whether the winner's score always exceeds \( 1-T \). If so, no runner-up can exceed \( T \) (since all class scores sum to 1), allowing the inner loop to be skipped. This check is empirically successful in most cases, bypassing the class-by-class inner loop.
	
	\paragraph{Break Optimization (\textcolor{blue}{blue} lines in \cref{alg:combined-break-then-continue-optimizations}).}
	A second improvement terminates verification early by checking whether the winner's score always exceeds the threshold at exit $k$. If this holds and no counterexample exists in prior exits, then all inputs advance to exit $k$ where the original winner consistently prevails---verification can \emph{break} and soundly return \safe{} without checking later exits.

	\begin{theorem}\label{the:optimized-verify_Pee}
		If the underlying verification tool is sound and complete, \cref{alg:combined-break-then-continue-optimizations} is sound and complete.
	\end{theorem}
	\begin{theorem}\label{the:optmized_verifies_Pee_comp}
		Given a network $\nn$ with EEs and ReLU activations, layer width bound $k$, input $x$, and $\epsilon > 0$, if $\tau(x)$ is \textit{stable} in $B_\epsilon^x$, then \cref{alg:combined-break-then-continue-optimizations} runtime is $\mathcal{O}(2^{k \cdot |\tau(x)|})\cdot poly(\text{\#neurons in }\nn)$.
	\end{theorem}
	
	To summarize: we formalized a robustness property for DNNs with EEs and proved it is fixed-parameter tractable (\cref{the:Pee_is_FPT}) under trace stability.
	We presented a sound and complete baseline algorithm (\cref{alg:basic-early-exit-verification}), then introduced break-and-continue heuristics (\cref{alg:combined-break-then-continue-optimizations}) and showed they yield a tighter complexity bound (\cref{the:optmized_verifies_Pee_comp}).
	Proofs of soundness and completeness (\cref{the:basic_verifies_Pee,the:optimized-verify_Pee}) and of complexity results (\cref{the:Pee_is_FPT,the:optmized_verifies_Pee_comp}) can be found in \cref{appendix:Proofs}.
	The framework extends naturally to other properties and activation functions (see \cref{app:extendability}).

	\section{Evaluation}
	\label{sec:results}
	
	We evaluate \pave{} across a diverse set of networks and datasets to answer four questions:
	\begin{inparaenum}[(i)]
		\item Does \pave{} work in practice and what are its limits?
		\item Does adding EEs improve verifiability compared to standard (no-EE) verification?
		\item Does trace stability hold empirically, explaining the efficiency gains?
		\item How do EE design choices (threshold, exit placement) affect the accuracy--robustness--verifiability trade-off?
	\end{inparaenum}
	
	\subsection{Benchmarks, Experimental Setup and Model Training}
	\label{sec:setup}
	We conducted experiments on MNIST~\citep{LeCoBu2010} and CIFAR-10/100~\citep{Kr2009}, with multiple architectures: Fully Connected, LeNet-5 (CNN)~\citep{LeBoBeHa1998}, ResNet-18~\citep{HeZhReSu2016}, and VGG-16~\citep{SiZi2015}. Full training details appear in \cref{app:benchmarks-technical-details}.
	
	We adopt the sequential EE training procedure from~\citep{TeMcKu16,WaCaTaJuKeFu2020}: a baseline model is first trained without exits; EEs (each a fully connected layer with SoftMax) are then added and trained sequentially, keeping the base model fixed. This ensures that EEs do not degrade the original model's accuracy.

	We place exits at the ends of computational blocks and select thresholds on a validation accuracy-latency Pareto curve; these choices preserve the backbone and provide an early-exit population while exposing the accuracy-latency-verifiability trade-off in \cref{subsec:eval-practicality}.
	
	We implemented our framework in PyTorch and ran experiments on an NVIDIA A100 GPU (80 GB HBM2e). As an underlying verification tool, our framework uses (but is not limited to) $\alpha$-$\beta$ CROWN~\citep{xu2020automatic,XuZhWaWaJaLiHs21, WaZhXuLiJaHsKo21}, a SOTA method for formally verifying DNN robustness properties. $\alpha$-$\beta$ CROWN does not currently support nested AND operators. We circumvent this by encoding each query as a collection of smaller queries, one per conjunct, and use the same encoding for the original queries for a fair comparison. More details in \cref{app:encoding-and-verifier-choice}.
	
	\cref{tab:model_metrics} summarizes the benchmarks. EE accuracy changes by 
	at most approximately 0.3\% from the base accuracy. The benchmarks span 
	600K to over 33M parameters, demonstrating acceleration on moderately 
	large, standard-trained image classifiers. 
	Our claim is the reduction of verification work for a fixed network 
	and specification: difficulty depends on the architecture, specification, 
	perturbed dimensions, radius, training, timeout, and backend, not 
	parameter count alone. For example, the 138M-parameter VGG-16 VNN-COMP 
	benchmark~\cite{BrBaJoWu24} perturbs only a small subset of ImageNet 
	pixels in some instances, whereas our CIFAR queries perturb all 3,072 
	pixels.
	
	\begin{table}[b]
		\centering
		\caption{Characteristics and evaluation metrics of the benchmarks. 
		}
		\label{tab:model_metrics}
		\resizebox{\textwidth}{!}{
			\begin{tabular}{c|c|c|c|c|c|c}
				\toprule
				\textbf{Model} & \textbf{Dataset} & \textbf{Size} & \textbf{\# Layers} & \textbf{Accuracy} & \textbf{EE Accuracy} & \textbf{Exit Distribution} \\
				\midrule\midrule
				{\textbf{FC-6}} & {MNIST} & {1,519,720} & {6} & {98.18} & {98.2} & {[9772, 152, 46, 30]} \\
				{\textbf{CNN (LeNet-5)}} & {CIFAR-10} & {596,178} & {5} & {70.02} & {69.93} & {[5924, 2066, 2010]} \\
				{\textbf{ResNet-18}} & {CIFAR-10} & {11,243,102} & {18} & {86.43} & {86.11} & {[6656, 1899, 1445]} \\
				{\textbf{VGG-16}} & {CIFAR-10} & {33,769,566} & {13} & {93.45} & {93.14} & {[8268, 1278, 454]} \\
				\bottomrule\bottomrule
			\end{tabular}
		}
	\end{table}
	
	\subsection{Evaluating the Practicality of Pareto Verification}\label{subsec:eval-practicality}
	We evaluate \pave{} (\cref{alg:combined-break-then-continue-optimizations}) against the standard Vanilla verifier (no EEs) on the four benchmarks of \cref{tab:model_metrics}. We used 100 samples per benchmark per $\epsilon$, with $\epsilon \in \{0.1, 0.05, 0.01, 0.005, 0.001\}$ and a 30-minute per-query timeout. Queries that hit the timeout or triggered an assertion failure in the underlying verifier are counted as \emph{unresolved}; note that unresolved outcomes come from the underlying verifier, not from our algorithm.
	
\begin{figure}[ht!]
		\centering
		\includegraphics[width=0.95\textwidth]{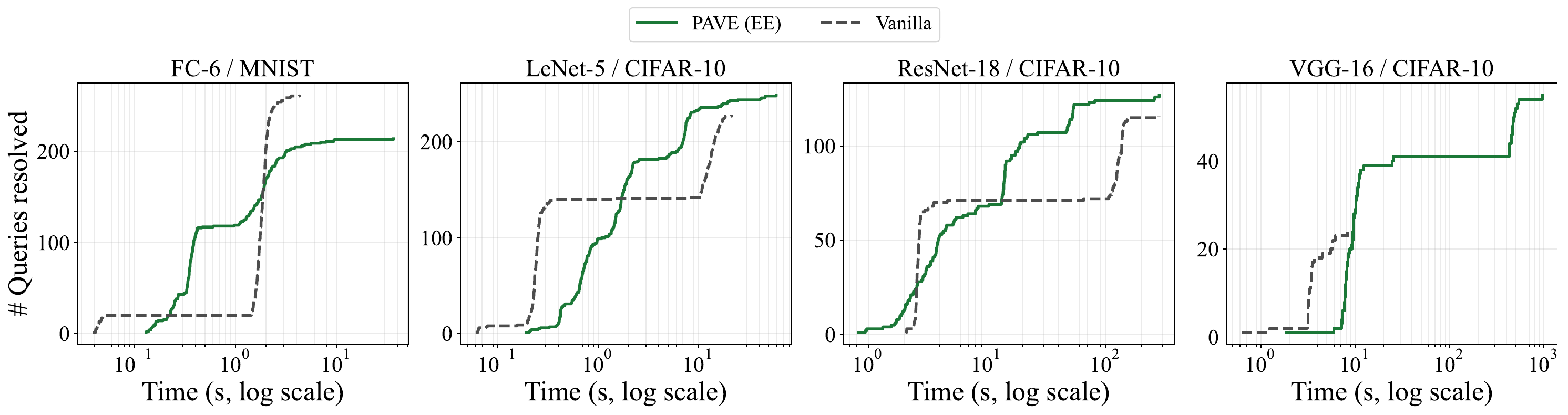}
		\caption{Number of queries resolved (\safe{}~+~\unsafe{}) within a given wall-clock budget, per benchmark. Higher and further-left curves are better. \pave{} (green) rises earlier than Vanilla (gray dashed) on every benchmark and plateaus higher on every CIFAR-10 network. The results show that the \pave{} benefit grows with network size---modest on the small FC-6 and largest on VGG-16.
		}
		\label{fig:cactus}
		
		\vspace{1em}
		\includegraphics[width=0.95\textwidth]{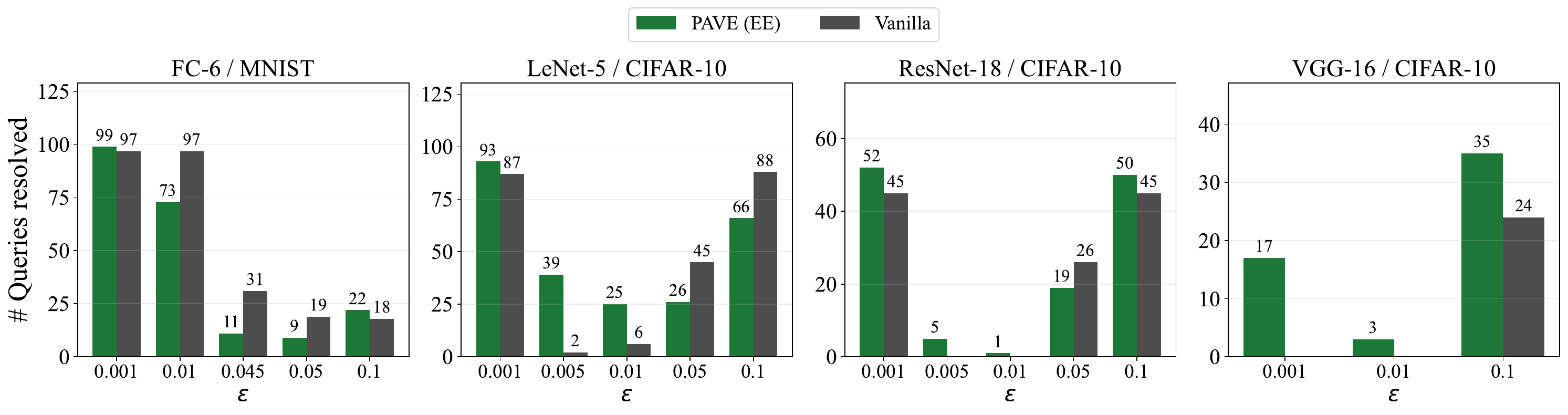}
		\caption{Number of queries resolved per~$\epsilon$, on a matched $\epsilon$ grid. Missing Vanilla bars indicate that Vanilla resolved \emph{zero} queries at that $\epsilon$ within the timeout (ResNet-18 at $\epsilon\in\{0.005, 0.01\}$, VGG-16 at $\epsilon\in\{0.001, 0.01\}$); \pave{} resolves queries in all of those cells. As in \cref{fig:cactus}, the benefit grows with network size: on FC-6, Vanilla matches or exceeds \pave{} coverage; on the larger CIFAR-10 networks, \pave{} adds coverage in the hard mid-range $\epsilon$.}
		\label{fig:resolved-counts}
	\end{figure}

	We report two complementary views of the data. \cref{fig:cactus} is a cactus / survival plot: for each benchmark, the number of queries each method resolves within a given wall-clock budget. \cref{fig:resolved-counts} breaks the same data down per~$\epsilon$: for each epsilon value, the number of queries each method resolves.
	
	\paragraph{\pave{} solves more queries, and earlier.}
	\cref{fig:cactus} makes two claims at once: on all three CIFAR-10 benchmarks \pave{} resolves more queries overall (higher plateau) and resolves each individual query faster on average (curve rises earlier). On VGG-16, \pave{} resolves more than twice as many queries as Vanilla (55 vs.\ 24)---Vanilla times out on most \safe{} queries, while \pave{} resolves them. On ResNet-18, \pave{} likewise resolves more queries than Vanilla, and on LeNet-5 it finishes the same set of queries substantially earlier. These are the benchmarks where verification cost actually matters, and they are exactly the regime where \pave{} delivers.
	
	\paragraph{\pave{} unlocks the hard mid-range $\epsilon$.}
	\cref{fig:resolved-counts} breaks coverage down by~$\epsilon$ and clarifies \emph{where} the gains come from. At very small~$\epsilon$ most queries are trivially \safe{} and both methods resolve them; at very large~$\epsilon$ most are trivially \unsafe{} and counterexamples are found quickly by both. The interesting regime is the mid-range, where queries are genuinely hard and the underlying verifier spends its budget. This is where \pave{} dominates: on LeNet-5 at $\epsilon=0.005$, \pave{} resolves 39 queries while Vanilla resolves only 2; on VGG-16 at $\epsilon\in\{0.001,0.01\}$, \pave{} resolves 17 and 3 queries while Vanilla resolves none; on ResNet-18 at $\epsilon\in\{0.005, 0.01\}$, \pave{} resolves 5 and 1 queries while Vanilla again resolves none. Missing Vanilla bars in \cref{fig:resolved-counts} correspond exactly to these zero-coverage cases. By sidestepping the full network for inputs that exit early, \pave{} makes tractable the queries that the underlying verifier simply cannot solve within the timeout.
	
	\paragraph{The \pave{} advantage grows with network size.}
	Both figures show a clear size-monotone pattern. On the smallest benchmark 
	(FC-6 / MNIST, $\approx1.5$M parameters), Vanilla verification already 
	works well and in fact resolves slightly more queries than \pave{} 
	overall---this is expected, since FC-6 is small enough that the underlying 
	verifier does not need help with coverage. As networks grow (LeNet-5, 
	ResNet-18, VGG-16), \pave{} overtakes Vanilla and the gap widens, reaching 
	its maximum on the largest network. This is the intended regime for 
	\pave{}: the larger and deeper the network, the more the underlying 
	verifier benefits from only having to reason about a shallow prefix. On 
	FC-6, the \pave{} advantage instead shows up as reduced runtime per query 
	(\cref{tab:verification-stats} and \cref{fig:vtime_vanilla_vs_ee-alg4}).
	
	\paragraph{Break and continue ablation.}
	\cref{tab:verification-stats} compares \cref{alg:basic-early-exit-verification} (no optimizations) against \cref{alg:combined-break-then-continue-optimizations} (with the break and continue optimizations). Differences on \unsafe{} cases are minor---counterexamples are found quickly regardless---but \safe{} cases improve by up to $10\times$. On VGG-16, \cref{alg:basic-early-exit-verification} could not prove any \safe{} case before timing out, while \cref{alg:combined-break-then-continue-optimizations} succeeded with a 1532-second mean verification time, making VGG-16 verification feasible.
	
	\begin{table}[t]
		\centering
		\caption{Verification runtime statistics (seconds) per benchmark.
			Rows correspond to \cref{alg:basic-early-exit-verification} (top) and \cref{alg:combined-break-then-continue-optimizations} (bottom).}
		\label{tab:verification-stats}
		\resizebox{0.7\textwidth}{!}{
			\begin{tabular}{c|c|c|c|c|c|c}
				\toprule
				\textbf{Benchmark}
				& \multicolumn{3}{c|}{{\sat}}
				& \multicolumn{3}{c}{{\unsat}} \\
				\cline{2-7}
				& {Mean} & {Std} & {Median}
				& {Mean} & {Std} & {Median} \\
				\midrule \midrule
				\multirow{2}{*}{MNIST, FC6}
				& 0.527 & 1.903 & 0.068 & 13.037 & 3.185 & 13.032 \\
				& 0.600 & 2.739 & 0.201 & 1.143  & 1.205 & 0.397\\
				\midrule
				\multirow{2}{*}{CIFAR10, LeNet}
				& 2.250 & 6.670 & 0.320 & 56.927 & 9.516 & 58.628 \\
				& 2.279 & 4.556 & 0.688 & 7.255  & 10.550 & 6.185 \\
				\midrule
				\multirow{2}{*}{CIFAR10, ResNet18}
				& 3.518 & 12.499 & 1.306 & 409.395 & 74.235 & 375.703 \\
				& 5.679 & 12.316 & 3.197 & 42.646  & 60.521 & 19.061  \\
				\midrule
				\multirow{2}{*}{CIFAR10, VGG16}
				& 5.412 & 1.032 & 5.588 & -- & -- & -- \\
				& 8.823 & 1.818 & 8.737 & 1532.418 & 348.150 & 1417.474 \\
				\bottomrule \bottomrule
			\end{tabular}
		}
	\end{table}
	
	\paragraph{Scaling to CIFAR-100.}
	We additionally verified a ResNet-18 model on CIFAR-100. 
	Both the baseline and the EE model identified \unsafe{} cases; 
	however, the EE model verified \safe{} examples in about one hour, 
	while the baseline reached a two-hour timeout. At $\epsilon = 0.001$, 
	14 samples were verified as \safe{} only with EEs, while all were 
	\unknown{} for the baseline---demonstrating that \pave{} can enable 
	verification of otherwise infeasible cases.

	\subsection{\pave{} vs. Classical DNN Verification}
	
	We compare verification times for networks with and without EEs to demonstrate the potential of \pave{} for improving any DNN's verifiability.

\begin{figure}[b]
		\centering
		\includegraphics[width=0.9\textwidth]{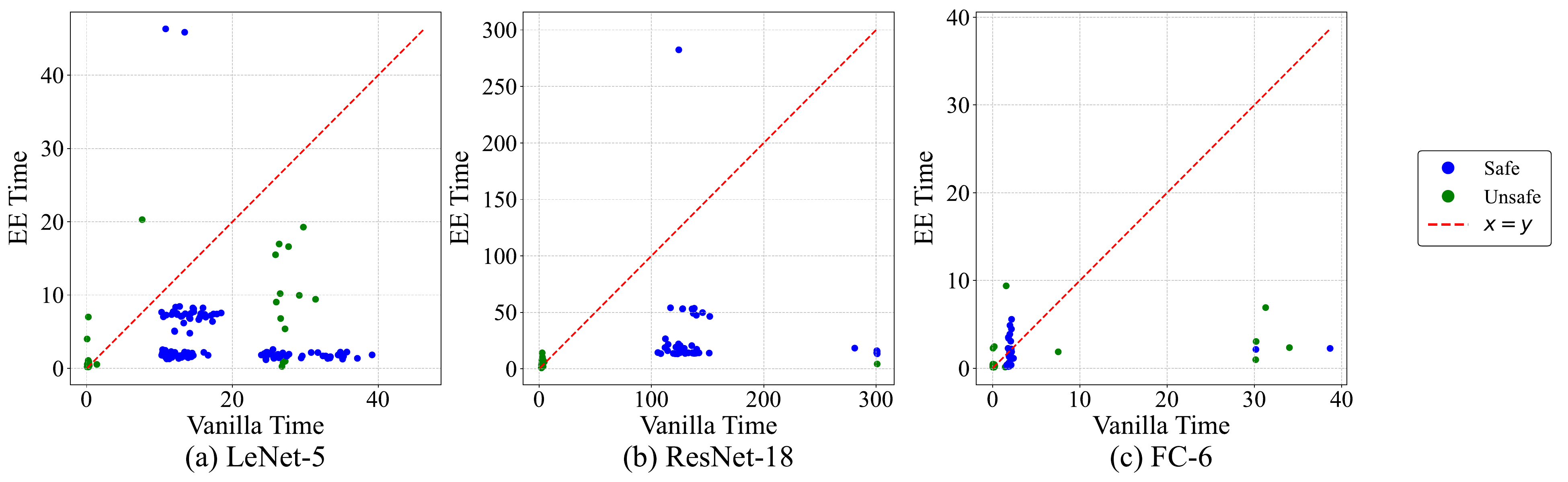}
		\caption{Comparison of verification times between the original model with the underlying verifier (Vanilla) and the model with EEs, using \cref{alg:combined-break-then-continue-optimizations}. Each point is a single query; the diagonal marks equal time.}
		\label{fig:vtime_vanilla_vs_ee-alg4}
	\end{figure}

	\cref{fig:vtime_vanilla_vs_ee-alg4} shows that EE-augmented DNNs are verified significantly more efficiently. While verification time for simple queries remains largely unchanged, harder queries benefit from substantial speedups.
	
\begin{figure}[t]
		\centering
		\begin{minipage}[b]{0.245\textwidth}
			\centering
			\includegraphics[width=0.9\textwidth]{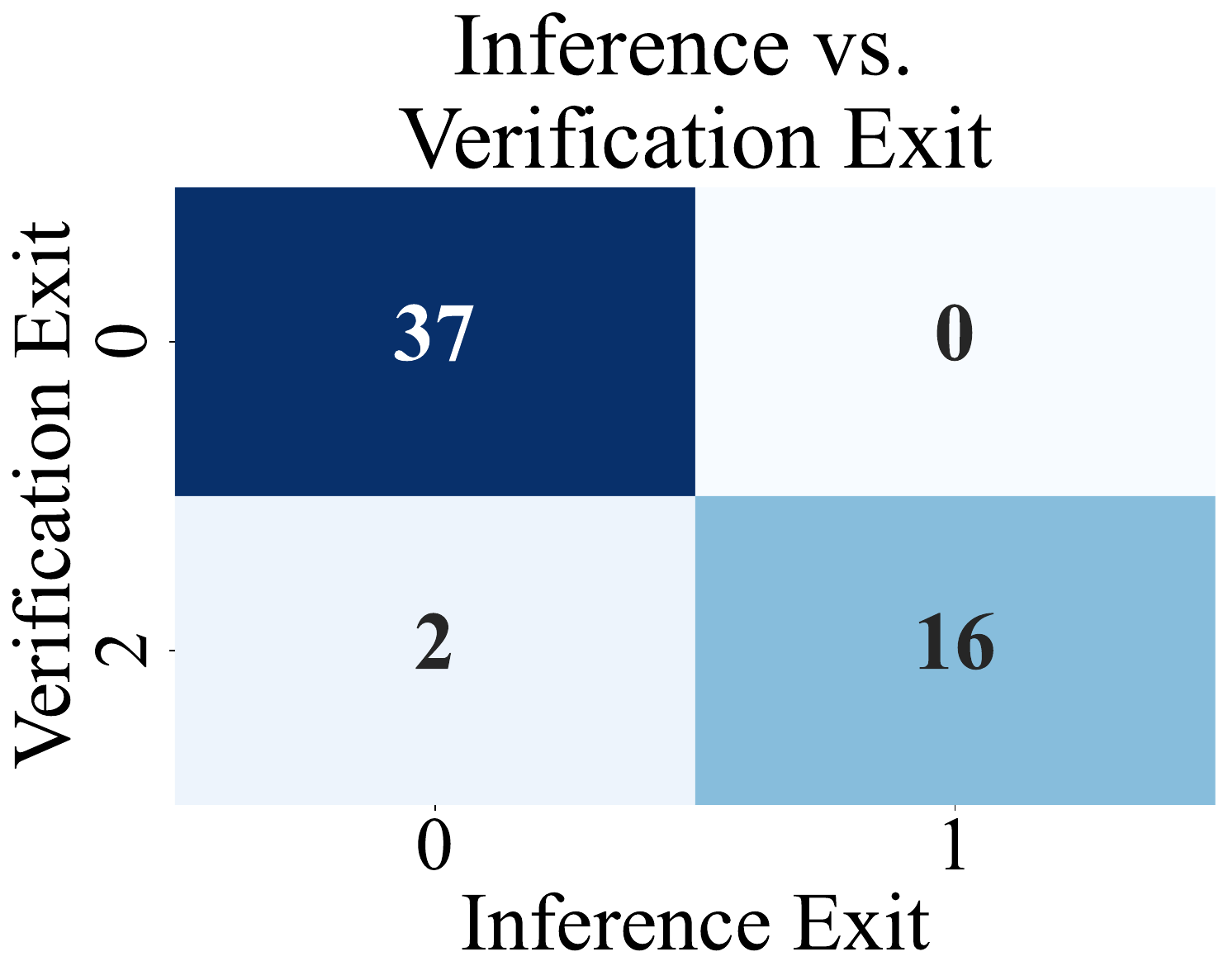}
			\caption*{\small{(a) ResNet-18 \safe{}}}
		\end{minipage}
		\hfill
		\begin{minipage}[b]{0.245\textwidth}
			\centering
			\includegraphics[width=0.9\textwidth]{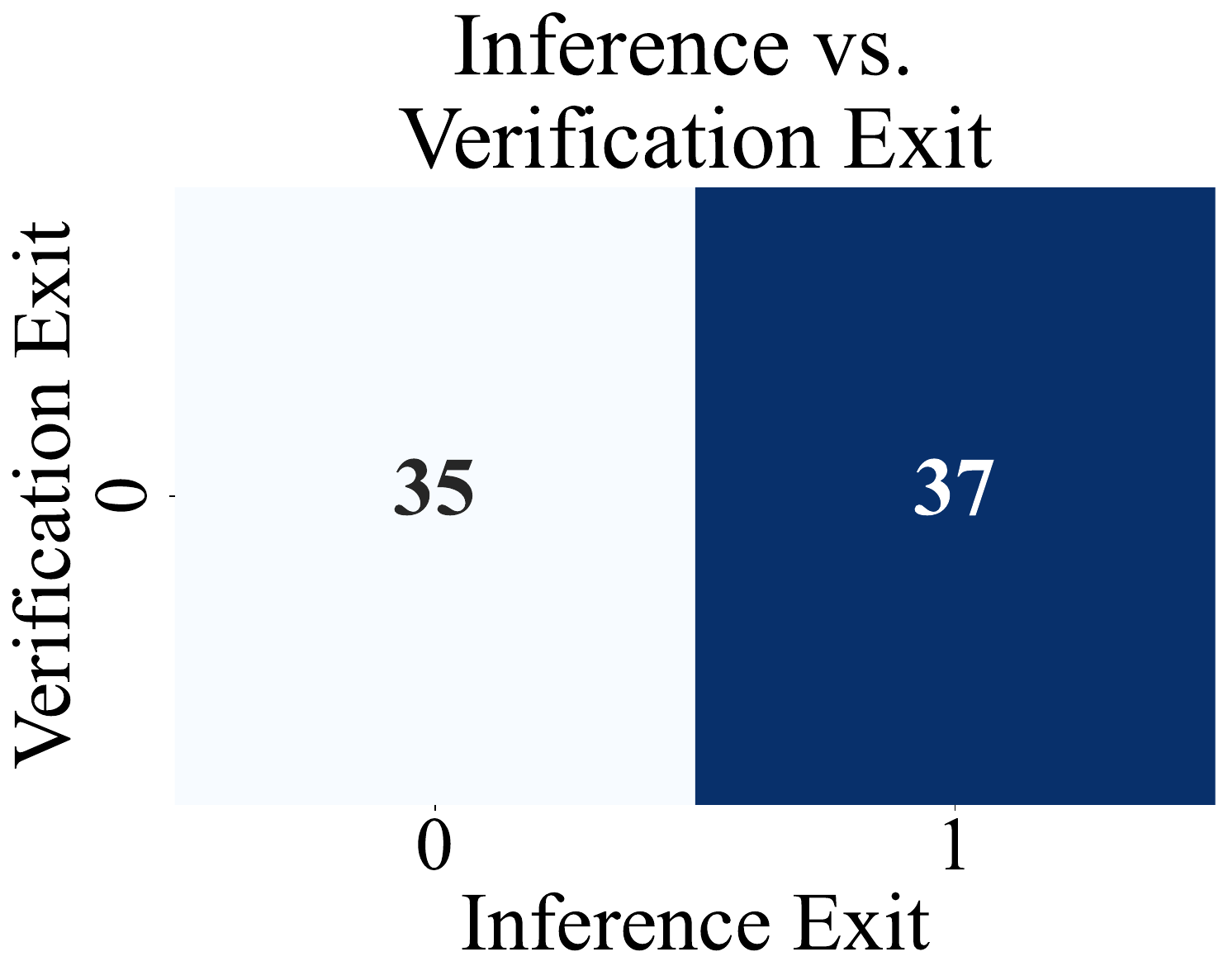}
			\caption*{\small{(b) ResNet-18 \unsafe{}}}
		\end{minipage}
		\hfill
		\begin{minipage}[b]{0.245\textwidth}
			\centering
			\includegraphics[width=0.9\textwidth]{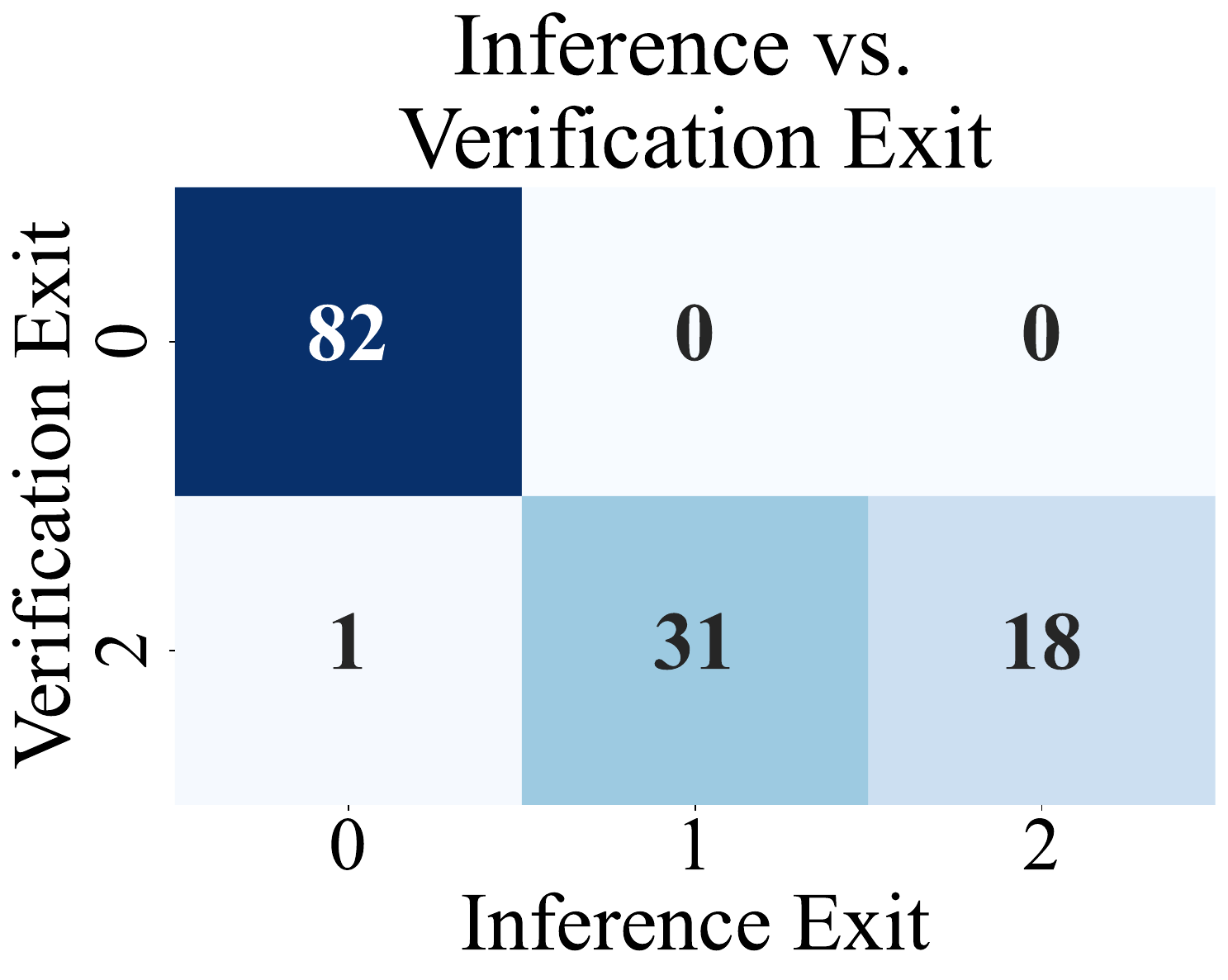}
			\caption*{\small{(c) LeNet-5 \safe{}}}
		\end{minipage}
		\hfill
		\begin{minipage}[b]{0.245\textwidth}
			\centering
			\includegraphics[width=0.9\textwidth]{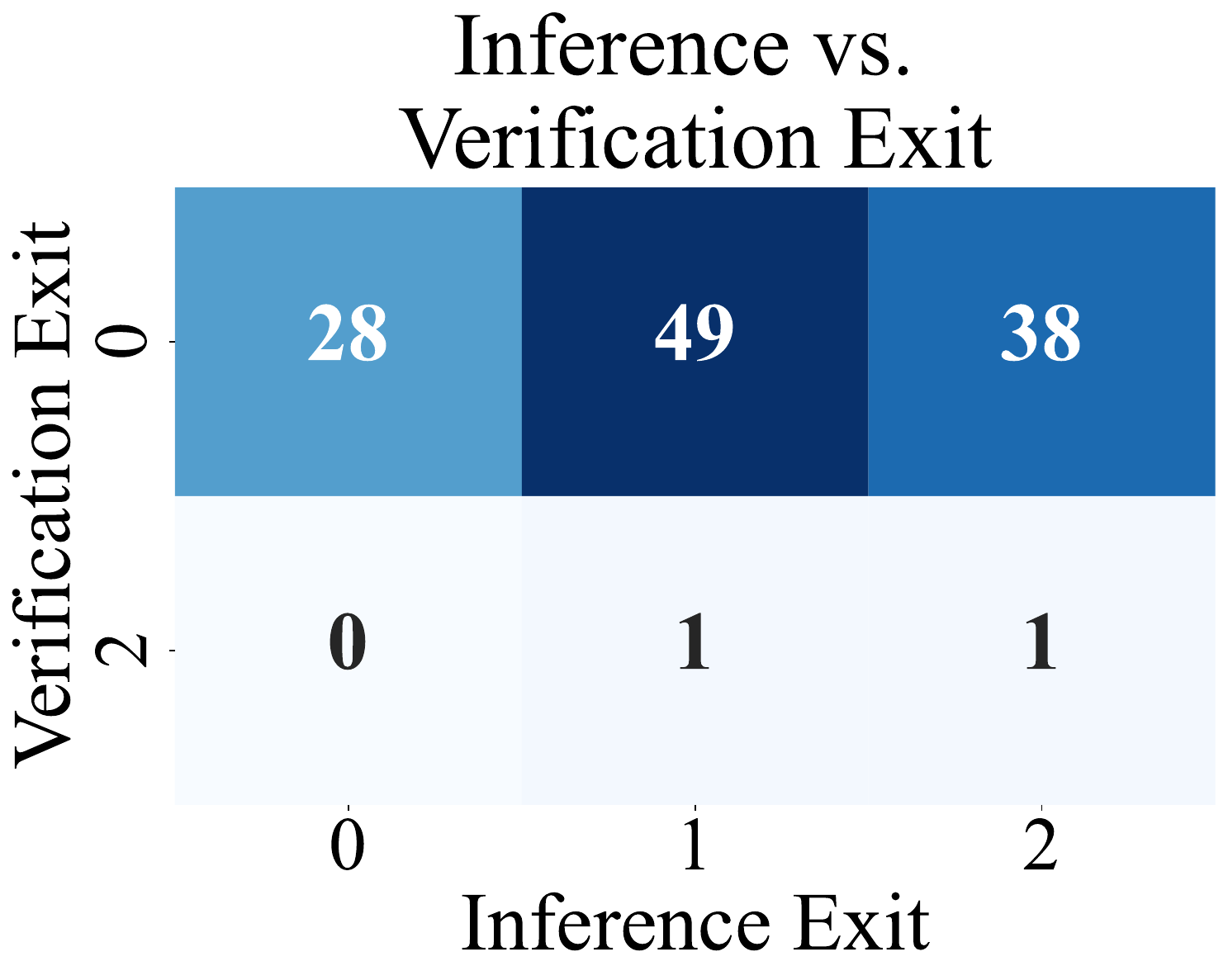}
			\caption*{\small{(d) LeNet-5 \unsafe{}}}
		\end{minipage}
		\caption{Heatmaps of inference vs.\ verification exit layer. Strong diagonal correlation for \safe{} cases confirms that trace stability holds in practice, explaining why \pave{} terminates early.}
		\label{fig:inf-vs-ver-exit-heatmap}
	\end{figure}

	To understand \emph{why} \pave{} is faster, \cref{fig:inf-vs-ver-exit-heatmap} compares inference exit layers with verification exit layers. A strong correlation emerges for \safe{} cases: inputs that exit at an early layer during inference also verify at that early layer, confirming that trace stability holds in practice. No such correlation appears for \unsafe{} cases (counterexamples behave differently from the original sample), which is expected; we discuss why trace stability cannot be expected to hold on \unsafe{} inputs in \cref{app:trace-stability-unsafe}. Additional results for MNIST are in \cref{app:verification_exit_vs_inference_exit}.
	
	\subsection{Robustness Analysis and Design Guidelines}
	Training networks with EEs entails a trade-off between predictive accuracy and inference latency. Different hyperparameter settings yield distinct working points, and users must select the configuration that best satisfies their constraints.
	
\begin{figure}[b]
		\centering
		\includegraphics[width=0.85\textwidth]{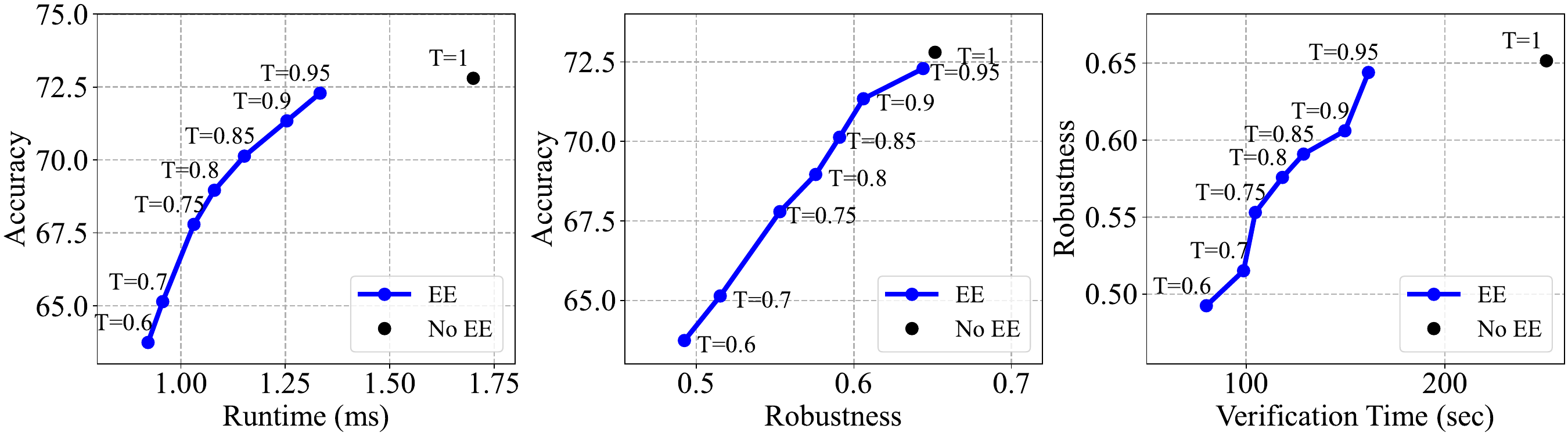}
		\caption{Impact of threshold selection on accuracy vs.\ runtime (left), accuracy vs.\ robustness (middle), and robustness vs.\ verification time (right) for CNN on CIFAR-10, $\epsilon = 0.005$.}
		\label{fig:acc-lat-rob}
	\end{figure}

	\paragraph{Impact of the Early Stop Threshold.}
	Adjusting the confidence threshold at each exit introduces a trade-off between accuracy and latency. \cref{fig:acc-lat-rob} (left) shows that higher thresholds improve accuracy but increase inference time. To quantify robustness, we measure it as the proportion of inputs formally verified as \safe{}, i.e.\ \#\safe{} / (\#\safe{} + \#\unsafe{}).
	With the EE architecture fixed and only the threshold varied, \cref{fig:acc-lat-rob} (middle) shows that robustness closely follows accuracy---higher thresholds boost both. However, \cref{fig:acc-lat-rob} (right) reveals that verification time also grows with the threshold. Consequently, threshold selection requires balancing accuracy, latency, robustness, and verifiability jointly.
	
\begin{figure}[t]
		\centering
		\begin{minipage}[b]{0.5\textwidth}
			\centering
			\includegraphics[width=\linewidth]{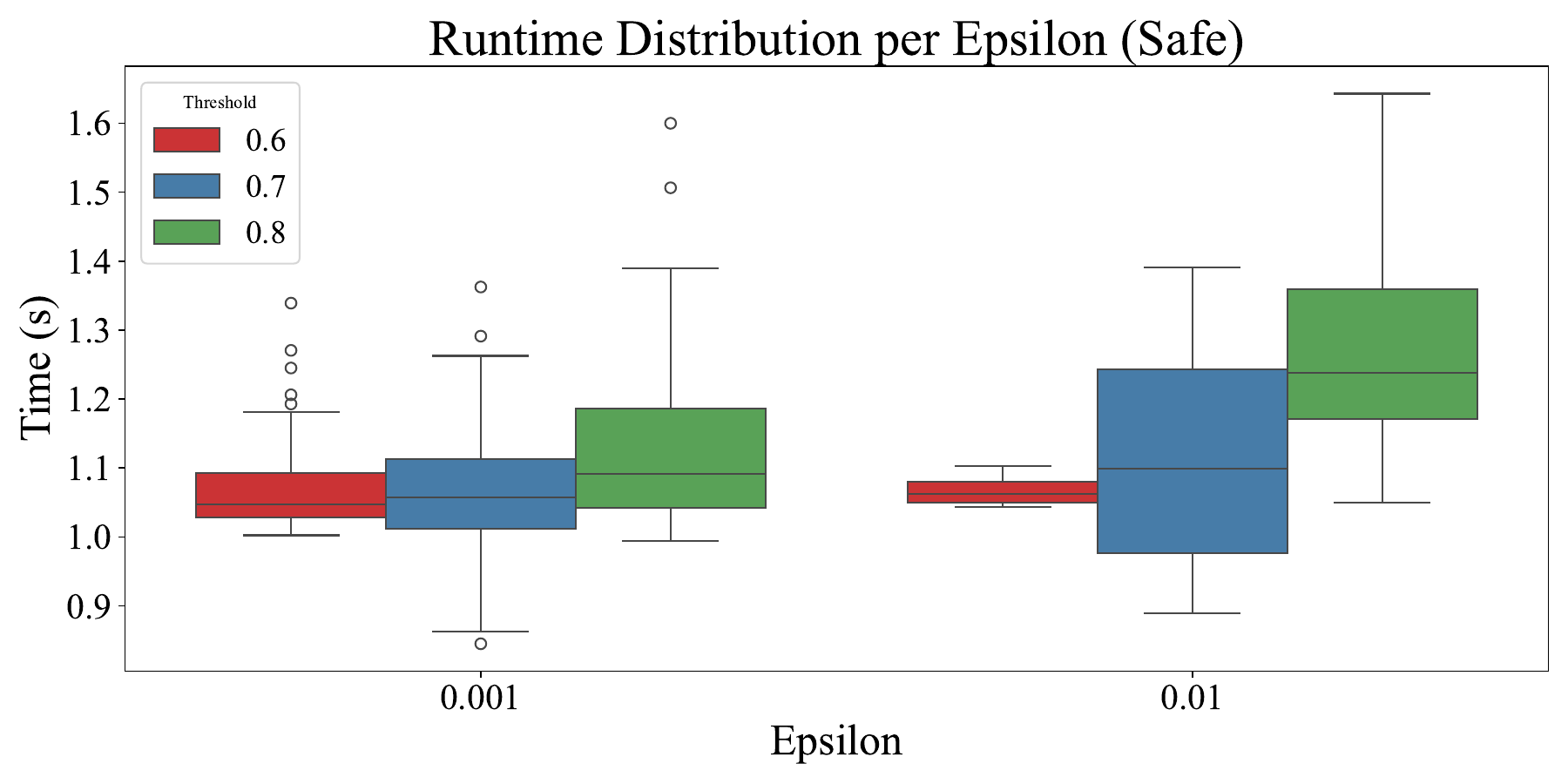}
			\caption{Impact of threshold on verification time for CNN on CIFAR-10.}
			\label{fig:bigger_T_implies_bigger_verification_time}
		\end{minipage}
		\hfill
		\begin{minipage}[b]{0.37\textwidth}
			\centering
			\includegraphics[width=\linewidth]{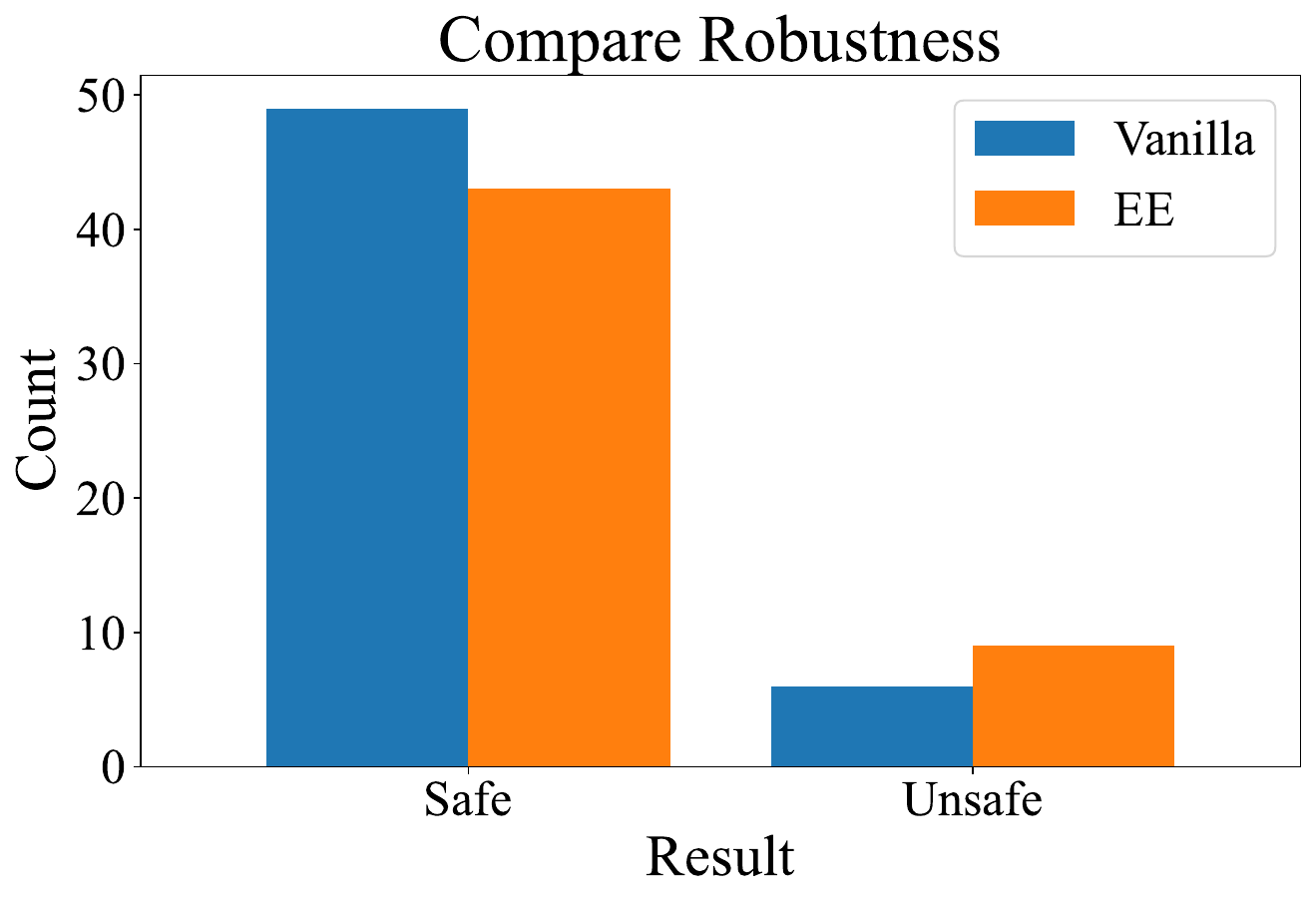}
			\caption{Robustness comparison between vanilla and EE-augmented models.}
			\label{fig:compare_robustness}
		\end{minipage}
	\end{figure}

	To dissect verification cost further, \cref{fig:bigger_T_implies_bigger_verification_time} breaks down verification times across $\epsilon$ values. Both mean and variance grow with the threshold, reinforcing that more conservative exit criteria demand heavier verification effort. \cref{fig:compare_robustness} compares against the vanilla network (no EEs, $T=1$): the vanilla model achieves the highest robustness but requires the longest verification time.
	
	\paragraph{Impact of Exit Location.}
	Fixing the threshold and varying the exit position, we compare two ResNet-18 variants: one with an exit after block 1 (\textbf{RN\_ee1}) and one after block 2 (\textbf{RN\_ee2}). \cref{tab:resnet-ee} shows that earlier exits increase robustness and verifiability (fewer \unknown{}, shorter verification time) at a minor accuracy cost; later exits preserve accuracy but reduce verifiability.
	
	\begin{table}[t]
		\centering
		\caption{Verification statistics for ResNet-18 variants with a single early exit.}
		\label{tab:resnet-ee}
		\resizebox{\textwidth}{!}{
			\begin{tabular}{c|c|c|ccc|cc}
				\toprule
				Architecture & Accuracy & Inference Time & {\#\safe{}} & {\#\unsafe{}} & {\#\unknown{}} & Safe Ver Time & Unsafe Ver Time \\
				\midrule
				\textbf{RN\_ee1} & 0.8757 & 30.56ms & 121 & 109 & 190 & 21.4s & 11.7s \\
				\textbf{RN\_ee2} & 0.8921 & 36.94ms &  85 & 108 & 227 & 23.4s & 14.2s \\
				\bottomrule\bottomrule
			\end{tabular}
		}
	\end{table}
	
	We find that earlier exits raise robustness and reduce verification time, while later exits trade verifiability for accuracy. These insights give practitioners actionable design guidance for choosing exit locations that best balance their accuracy, latency, robustness, and safety requirements.

	\section{Related Work}\label{sec:related_work}
	
	\paragraph{Formal Verification of DNNs.}
	The formal verification of DNNs has received growing attention~\citep{BrMuBaJoLi23,BrBaJoWu24} due to their increasing use in safety-critical domains. Early efforts focused on fully-connected and convolutional networks~\citep{KaBaDiJuKo17,GeMiDrTsChVe18,BuTuToKuLuKo20}, while more recent work has expanded to RNNs~\citep{KhNeRoBaBoFiHaLeYe20}, LSTMs~\citep{MoFiFr23}, transformers~\citep{ShZhChHuHs20}, and GNNs~\citep{WuBaShNaSi22,LaEiAl25}.
	
	
	Various techniques have been proposed to improve scalability, including symbolic propagation, abstract interpretation, abstraction-refinement, adversarial pruning, certified training, and training for verification ~\citep{WaPeWhYaJa18,WaPeWhYaJa18_2,SiGePuVe19,WaZhXuLiJaHsKo21,ElGoKa20,ElCoKa22,CoElBaKa23,XuZhWaWaJaLiHs21,ZhWaXuLiLiJaHsKo22,ZhBrHaZh24,MuEcFiVe23,DeBuDvKuStLo24,MaMuFiVe23,DuXuNgDy24,LiZhSoSuYaHuZh25}. \pave{}, which adds lightweight EE heads post-hoc, is \emph{orthogonal} to these techniques: improved verifiers or architectures directly improve \pave{} too.
	
	
	
	\paragraph{Early Exit Architectures.}
	EE represents a dynamic inference strategy within the broader landscape of DNN optimization, which also includes quantization~\citep{ChWaZhZh17}, pruning~\citep{FrCa19}, knowledge distillation~\citep{PhLa19}, and neural architecture search~\citep{BaQu17}. Models with EEs accelerate inference by allowing early termination for easy inputs~\citep{RaSrChPaCo24,DiMa25,DiXuHaMe24,SaDiBr2022}. Our work is, to our knowledge, the first to explore how EEs can be exploited for scalable formal verification of neural networks.
	

	\section{Conclusion and Future Work}\label{sec:conclusion}
	
	\paragraph{Conclusion.}
	We introduced \pave{}, a paradigm that exploits EE architectures to solve the large majority of robustness queries in a fraction of the time by decomposing hard DNN verification problems into tractable sub-problems. We showed that EEs can be added to many standard DNNs without accuracy degradation, formalized the EE robustness property and proved its FPT complexity, and proposed a sound and complete algorithm with two optimizations that achieve up to $10\times$ speedup on \safe{} cases. Experiments demonstrate that \pave{} enables verification of queries that are otherwise intractable, and provides actionable design guidance on EE hyperparameters.

	\paragraph{Future Work.}
	Several directions remain open, including evaluation on larger and non-image models, extension to other properties, distributed verification across exits, and alternative exit conditions~\citep{RaSrChPaCo24}.

	\section{Acknowledgements}
    The work of Elboher and Katz was partially funded by the European Union (RobustifAI project, ID 101212818). Views and opinions expressed are however those of the author(s) only and do not necessarily reflect those of the European Union or the European Health and Digital Executive Agency (HADEA). Neither the European Union nor the granting authority can be held responsible for them.
    The work of Raviv and Kugler was supported by the Horizon 2020 research and innovation programme for the Bio4Comp project under grant agreement number 732482, by the Israel Science Foundation (Grant No. 190/19) and by the Data Science Institute at Bar-Ilan University (seed grant).
	
	\bibliographystyle{plainnat}
	\bibliography{NeurIPS_2026_bib_v2}

	
	\newpage
	\appendix
	
	\section{An Example of a Neural Network with Early Exits}
	\label{app:early-exit-example-figure}
	
	\begin{figure}[ht!]
		\begin{center}
			\scalebox{0.45}{
				\def\layersep{2.5cm}
				\def\vertSepFactory{0.7}
				\begin{tikzpicture}[shorten >=1pt,->,draw=black!50, node distance=\layersep]
					\foreach \name / \y in {1,...,3}
					\node[input neuron, pin=left:Input \#\y] (I-\name) at (0,-\vertSepFactory * \y) {};
					
					\foreach \name / \y in {1,...,6}
					\path[yshift=1.4cm]
					node[hidden neuron] (H1-\name) at (1*\layersep,-\vertSepFactory * \y cm) {};
					
					\foreach \name / \y in {1,...,2}
					\node[output neuron, pin=right:Exit 1 \#\y]
					(EE1-\name) at (1.5*\layersep + 20, -\vertSepFactory * \y cm + 90) {};
					\foreach \source in {1,...,6}
					\foreach \dest in {1,...,2}
					\path (H1-\source) edge (EE1-\dest);
					
					\foreach \name / \y in {1,...,6}
					\path[yshift=1.4cm]
					node[hidden neuron] (H2-\name) at (2*\layersep,-\vertSepFactory * \y cm) {};
					
					\foreach \name / \y in {1,...,6}
					\path[yshift=1.4cm]
					node[hidden neuron] (H3-\name) at (3*\layersep,-\vertSepFactory * \y cm) {};
					
					\foreach \name / \y in {1,...,2}
					\node[output neuron, pin=right:Exit 2 \#\y]
					(EE2-\name) at (3.5*\layersep + 20, -\vertSepFactory * \y cm + 90) {};
					\foreach \source in {1,...,6}
					\foreach \dest in {1,...,2}
					\path (H3-\source) edge (EE2-\dest);
					
					\foreach \name / \y in {1,...,6}
					\path[yshift=1.4cm]
					node[hidden neuron] (H4-\name) at (4*\layersep,-\vertSepFactory * \y cm) {};
					
					\foreach \name / \y in {1,...,2}
					\node[output neuron,pin={[pin edge={->}]right:Output \#\y}]
					(O-\name) at (5*\layersep, -\vertSepFactory * \y cm) {};
					
					\foreach \source in {1,...,3}
					\foreach \dest in {1,...,6}
					\path (I-\source) edge (H1-\dest);
					
					\foreach \source in {1,...,6}
					\foreach \dest in {1,...,6}
					\path (H1-\source) edge (H2-\dest);
					
					\foreach \source in {1,...,6}
					\foreach \dest in {1,...,6}
					\path (H2-\source) edge (H3-\dest);
					
					\foreach \source in {1,...,6}
					\foreach \dest in {1,...,6}
					\path (H3-\source) edge (H4-\dest);
					
					\foreach \source in {1,...,6}
					\foreach \dest in {1,...,2}
					\path (H4-\source) edge (O-\dest);
				\end{tikzpicture}
			}
			\caption{A fully connected DNN with two EEs placed at the first and third layers. The neuron values at these exits are the results of applying SoftMax to the hidden values in the corresponding layers.}
			\label{fig:fullyConnectedNetworkWithExitsAbove}
		\end{center}
	\end{figure}

	\section{Complexity Analysis}
	\subsection{\cref{alg:basic-early-exit-verification}'s Complexity}
	We analyze the complexity of the proposed verification algorithms. Solving verification queries is NP-Hard~\citep{KaBaDiJuKo17}, and current methods have exponential complexity in the network's number of neurons in a worst-case scenario. We denote the worst-case complexity of the underlying verification tool by $\mathcal{O}(2^{N})$ where $N$ is the number of neurons in $\nn$.
	
	\begin{theorem}
		The worst-case complexity of \cref{alg:basic-early-exit-verification} is $\mathcal{O}(2^{N})$.
	\end{theorem}
	\begin{proof}
		\cref{alg:basic-early-exit-verification} applies $E\cdot(C-1)+1$ verification queries, where $E$ is the number of EEs and $C$ is the number of classes ($+1$ for the last query). Each query is $\mathcal{O}(2^{N_i})$ where $N_i$ is the number of neurons in the partial network until the $i$'th exit (including), summing to an overall $\mathcal{O}(2^{N})$ complexity.
	\end{proof}
	
	\subsection{\cref{alg:combined-break-then-continue-optimizations}'s Complexity}\label{app:optmized_verifies_Pee_comp_proof}
	\cref{alg:combined-break-then-continue-optimizations} performs break and continue heuristic optimizations, which are not necessarily triggered, and hence its worst-case complexity is similar to that of \cref{alg:basic-early-exit-verification}. However, its complexity decreases under the trace stability assumption. We prove \cref{the:optmized_verifies_Pee_comp}:
	\begin{theorem*}
		Given a network $\nn$ with EEs and ReLU activations, layer width bound $k$, input $x$, and $\epsilon > 0$, if $\tau(x)$ is \textit{stable} in $B_\epsilon^x$, then \cref{alg:combined-break-then-continue-optimizations} runtime is $\mathcal{O}(2^{k \cdot |\tau(x)|})\cdot poly(\text{\#neurons in }\nn)$.
	\end{theorem*}
	\begin{proof}
		We denote the index of the early exit where the output of $x$ is returned by $E_x$.
		In \cref{alg:combined-break-then-continue-optimizations}, line 8 checks if the winner always wins in the current exit. If the trace of each input in $B_\epsilon^x$ equals $\tau(x)$ and the result is \safe{}, it must be returned in line 8 in the $E_x$'th iteration: it cannot be returned before since $x$ itself constitutes a counterexample, and it is returned at iteration $E_x$ since the trace of all inputs equals $\tau(x)$. If the answer is \unsafe{}, the counterexample must be found by the $E_x$'th iteration; in each iteration, line 14 launches $ExistsPrevCEX(\nn,x,\epsilon,k)$, which checks all possible counterexamples until the $E_x$'th exit. Because $\tau(x')=\tau(x)$ for all inputs, all possible behaviors are examined after the $E_x$'th iteration.
		
		This means that \cref{alg:combined-break-then-continue-optimizations} will not proceed to iterations that run verification queries on sub-networks larger than the sub-network that $x$ was propagated through. The number of neurons in each such sub-network is no more than $k\cdot|\tau(x)|$, and solving verification queries of that size takes $2^{k\cdot|\tau(x)|}\cdot poly(\text{\#neurons in }\nn)$. Therefore, the complexity is bounded by $\mathcal{O}(2^{k\cdot|\tau(x)|})\cdot poly(\text{\#neurons in }\nn)$.
	\end{proof}

	\section{Proofs}
	\label{appendix:Proofs}
	
	In this appendix we provide the detailed proofs for the algorithms.
	
	\subsection{Proof of \cref{the:basic_verifies_Pee}}
	\label{app:basic_verifies_Pee_proof}
	\begin{theorem*}
		If the underlying verification tool is sound and complete, \cref{alg:basic-early-exit-verification} is sound and complete.
	\end{theorem*}
	\begin{proof}
		We split the proof for soundness into 3 parts:
		\begin{enumerate}
			\item{There is a satisfying example to $P_{ee}$ if and only if there are exit $k\in ee \cup \{last\}$ and runner-up $i\in\mathcal{C}\setminus\{w\}$ such that the runner-up wins at exit $k$ and there is no preceding layer where the winner wins. This is encoded in line 5 (for early exits) and line 7 (for the last exit) in \cref{alg:basic-early-exit-verification}.}
			\item{If \cref{alg:basic-early-exit-verification} returns $\sat$, one of the properties in lines 5 or 7 is $\sat$. The result is sound if the underlying verification tool is sound.}
			\item{Otherwise, if \cref{alg:basic-early-exit-verification} does not return $\sat$, it returns $\unsat$ at the last line. Since the algorithm iterates over all exits and all runner-ups, completeness of the underlying tool implies that if there is no counterexample then $P_{ee}$ is not satisfiable.}
		\end{enumerate}
		Completeness follows from the fact that there are finitely many queries, each guaranteed to terminate by completeness of the underlying tool.
	\end{proof}
	
	\subsection{Proof of \cref{the:Pee_is_FPT}}\label{app:Pee_is_FPT_proof}
	
	\begin{theorem*}
		Given a network $\nn$ with EEs and ReLU activations, layer width bound $k$, input $x$, and $\epsilon > 0$, if $\tau(x)$ is \textit{stable} in $B_\epsilon^x$, then solving $P_{ee}$ with $(\nn,x,\epsilon)$ is \textit{FPT}$(k \cdot |\tau(x)|)$.
	\end{theorem*}
	\begin{proof}
		Given that $\tau(x)$ is stable in $B_\epsilon^x$, every input in $B_\epsilon^x$ exits at the same exit as $\nn(x)$. The verification process therefore only needs to consider the layers up to that exit. The number of neurons up to this exit is bounded by $k\cdot|\tau(x)|$. Solving the query is done by splitting each ReLU into two cases---active $(y=x)$ and inactive $(y=0)$---giving $2^{k\cdot |\tau(x)|}$ combinations, each solvable in $poly(\text{\#neurons in }\nn)$ via linear programming~\citep{kh80}. Therefore, the problem is in \textit{FPT}$(k\cdot |\tau(x)|)$.
	\end{proof}
	
	\subsection{Proof of \cref{the:optimized-verify_Pee}}
	\label{app:optmized_verifies_Pee_proof}
	We separate the break and continue optimizations into \cref{alg:break-optimization} (blue lines only) and \cref{alg:continue-optimization} (orange lines only), prove soundness and completeness for each by equivalence to \cref{alg:basic-early-exit-verification}, and then derive the correctness of \cref{the:optimized-verify_Pee}.
	
	\begin{figure}[t]
		\centering
		\scalebox{0.9}{%
			\begin{minipage}{\columnwidth}
				\refstepcounter{algorithm}\label{alg:break-optimization}%
				\hrule height 0.8pt
				\vspace{2pt}
				\noindent\textbf{Algorithm \thealgorithm}\quad Pareto Verification -- Break Optimization
				\vspace{2pt}
				\hrule
				\vspace{4pt}
				\noindent\textbf{Input} $\nn$, $x$, $\epsilon_p$ \quad
				\textbf{Output} $\nn$ is robust in $B_{\epsilon}(x)$, or counterexample
				\begin{algorithmic}[1]
					\STATE {$w = argmax(\nn(x))$}
					\STATE{$ee$ := indices of layers with early exits in $\nn$}
					\STATE{last := index of last layer in $\nn$}
					\FOR {$k \in ee\cup \{last\}$}
					\IF{\textcolor{blue}{k $\neq$ last}}
					\STATE{\textcolor{blue}{$\mathcal{P}:=\exists x'\in B_{\epsilon}^x:\nn(x')^k_w<T$}}
					\ELSE
					\STATE{\textcolor{blue}{$\mathcal{P}:=\exists
							x'\in B_{\epsilon}^x,\exists i\in\mathcal{C}\setminus\{w\}: \nn(x')_w<\nn(x')_i
							$}}
					\ENDIF
					\STATE {\textcolor{blue}{res, cex = Verify($\nn, B^x_\epsilon,\mathcal{P}$)}}
					\IF{\textcolor{blue}{res == $\unsat$}}
					\RETURN {\textcolor{blue}{$\unsat$}}
					\ENDIF
					\STATE res, cex = $ExistsPrevCEX(\nn,x,\epsilon,k)$
					\IF {res == $\sat$}
					\RETURN {$\sat$, cex}
					\ENDIF
					\ENDFOR
					\RETURN $\unsat$
				\end{algorithmic}
				\vspace{4pt}
				\hrule height 0.8pt
			\end{minipage}%
		}%
	\end{figure}
	
	
	\begin{figure}[t]
		\centering
		\scalebox{0.9}{%
			\begin{minipage}{\columnwidth}
				\refstepcounter{algorithm}\label{alg:continue-optimization}%
				\hrule height 0.8pt
				\vspace{2pt}
				\noindent\textbf{Algorithm \thealgorithm}\quad Pareto Verification -- Continue Optimization
				\vspace{2pt}
				\hrule
				\vspace{4pt}
				\noindent\textbf{Input} $\nn$, $x$, $\epsilon_p$ \quad
				\textbf{Output} $\nn$ is robust in $B_{\epsilon}(x)$, or counterexample
				\begin{algorithmic}[1]
					\STATE {$w = argmax(\nn(x))$}
					\STATE{$ee$ := indices of layers with early exits in $\nn$}
					\STATE{last := index of last layer in $\nn$}
					\FOR {$k \in ee\cup \{last\}$}
					\STATE {\textcolor{orange}{$res$, $cex$ = Verify($\nn,B^x_\epsilon,\exists x'\in B^x_\epsilon: \nn^k_w(x')<1-T$)}}
					\IF {\textcolor{orange}{k == last $\lor$ res == $\sat$}}
					\STATE {res, cex = $ExistsPrevCEX(\nn,x,\epsilon,k)$}
					\IF {res == $\sat$}
					\RETURN {$\sat$, cex}
					\ENDIF
					\ENDIF
					\ENDFOR
					\RETURN $\unsat$
				\end{algorithmic}
				\vspace{4pt}
				\hrule height 0.8pt
			\end{minipage}%
		}%
	\end{figure}

	\begin{theorem}\label{the:continue-verifies_Pee}
		\cref{alg:continue-optimization} is equivalent to \cref{alg:basic-early-exit-verification}.
	\end{theorem}
	\begin{proof}
		The additional logic in \cref{alg:continue-optimization} (line 5) checks whether the winner's value might be smaller than \( 1-T \). If not, no runner-up can exceed \( T \) (scores sum to 1), so the result of $ExistsPrevCEX$ must be $\unsat$ and the inner loop can be skipped. If the condition is satisfied, or at the last layer (where the condition gives no assurance about runner-ups), the verification proceeds as in \cref{alg:basic-early-exit-verification}. In both cases the result is equivalent.
	\end{proof}
	
	\begin{theorem}\label{the:break-verifies_Pee}
		\cref{alg:break-optimization} is equivalent to \cref{alg:basic-early-exit-verification}.
	\end{theorem}
	\begin{proof}
		\cref{alg:break-optimization} adds a check at each iteration (line 10) for whether the winner might not win in the current exit. If $\unsat$ is returned (line 11 holds), verification halts and returns $\unsat$. This is valid because for all subsequent exits, neither line 5 nor line 7 of \cref{alg:basic-early-exit-verification} would hold---since there is a preceding exit where the winner wins for every example. If the condition is satisfiable, \cref{alg:break-optimization} continues to check for runner-up counterexamples, exactly as \cref{alg:basic-early-exit-verification} does, yielding equivalent results.
	\end{proof}
	
	\begin{theorem*}
		\cref{alg:combined-break-then-continue-optimizations} is equivalent to \cref{alg:basic-early-exit-verification}.
	\end{theorem*}
	\begin{proof}
		\cref{alg:combined-break-then-continue-optimizations} sequentially incorporates both \cref{alg:continue-optimization} and \cref{alg:break-optimization}. Correctness follows from \cref{the:continue-verifies_Pee} and \cref{the:break-verifies_Pee}.
	\end{proof}

	\section{Datasets and Models Technical Details}\label{app:benchmarks-technical-details}
	Below we describe each dataset and architecture, summarize key characteristics in~\cref{tab:benchmark_metadata}, and present the training protocol.
	
	We used three datasets:
	\begin{itemize}
		\item \textbf{MNIST}~\citep{LeCoBu2010}: Grayscale handwritten digits, 10 classes.
		\item \textbf{CIFAR-10}~\citep{Kr2009}: Color images, 10 classes; a standard benchmark for formal verification.
		\item \textbf{CIFAR-100}~\citep{Kr2009}: Extension of CIFAR-10 with 100 classes.
	\end{itemize}
	
	\begin{table}[h!]
		\centering
		\caption{Metadata for datasets used in the evaluation.}
		\label{tab:benchmark_metadata}
		\begin{tabular}{c|c|c|c|c}
				\toprule
				\textbf{Dataset} & \textbf{Train Size} & \textbf{Test Size} & \textbf{\#Classes} & \textbf{Input Shape} \\
				\midrule\midrule
				{\textbf{MNIST}} & {60,000} & {10,000} & {10} & {$1\times28\times28$} \\
				{\textbf{CIFAR-10}} & {50,000} & {10,000} & {10} & {$3\times32\times32$} \\
				{\textbf{CIFAR-100}} & {50,000} & {10,000} & {100} & {$3\times32\times32$} \\
				\bottomrule \bottomrule
		\end{tabular}
	\end{table}
	
	For each dataset, we trained models with EEs:
	\begin{itemize}
		\item \textbf{FC-6}: A fully connected architecture with 6 layers; first three equipped with EEs. Trained on MNIST.
		\item \textbf{LeNet-5 (CNN)}~\citep{LeBoBeHa1998}: 2 convolutional + 3 linear layers; EEs after first and second convolutional layers. Trained on CIFAR-10.
		\item \textbf{Modified ResNet-18}~\citep{HeZhReSu2016}: ResNet-18 with MaxPool replaced by AveragePool; EEs after the first and second blocks. Trained on CIFAR-10.
		\item \textbf{VGG-16}~\citep{SiZi2015}: Standard VGG-16 (13 convolutional layers, with Adaptive Average Pool instead of MaxPool for verifier compatibility); EEs after the 6th and 10th convolutional layers. Trained on CIFAR-10.
	\end{itemize}
	
	\paragraph{Training Protocol.}
	All models are trained using standard supervised learning. For CIFAR-10/100, we apply data augmentation (random cropping, horizontal flipping) and normalize with dataset-specific statistics.
	\textbf{MNIST} models are trained for 10 epochs (SGD, learning rate reduced $\times 0.1$ every 4 epochs).
	\textbf{CIFAR-10 models} (ResNet-18, VGG-16): 200 epochs, SGD with momentum 0.9 and weight decay $5\times10^{-4}$, step LR schedule at epochs 100 and 150. \textbf{LeNet-5}: 30 epochs, Adam, initial LR 0.001, decayed at epochs 10 and 20.
	\textbf{CIFAR-100 models}: Adam, LR 0.001, cosine annealing over 200 epochs.
	For EE heads: freeze backbone, train each exit sequentially for 20 epochs (10 for FC and LeNet) with the same optimizer and two $\times0.1$ LR decays.
	
	\paragraph{Technical Non-Triviality.}
	While conceptually simple, our method has two non-trivial aspects. First, it introduces many additional verification queries in the worst case, and its effectiveness depends on whether early exits succeed in verification; performance gains are not guaranteed and are validated only empirically. Second, each partial verification query verifies a more complex property involving the exit condition with Softmax, making the problem strictly harder. The benefit arises only when combined with a state-of-the-art verifier supporting Softmax~\citep{XuZhWaWaJaLiHs21,WaZhXuLiJaHsKo21}.
	
	\paragraph{Relation to Certified Training.}
	Unlike certified training methods, which modify the training process to improve robustness, our method preserves standard training and augments the model with EEs to accelerate verification. The two approaches are orthogonal and can be combined. For comparison: certified CNN-7 with $\epsilon \approx 0.0078$ achieves accuracy $78.82\%$ and robustness $64.41\%$~\citep{MaBaVe25}; our EE CNN-5 with $\epsilon = 0.005$ and $T=0.9$ achieves accuracy $71.34\%$ and robustness $60.61\%$ with a smaller network, demonstrating that our fine-tuning preserves robustness while providing efficiency gains.

	\section{Verification Exit Versus Inference Exit: MNIST \& FC-6}
	\label{app:verification_exit_vs_inference_exit}
	
	\begin{figure}[htbp]
		\centering
		\begin{minipage}[b]{0.48\textwidth}
			\centering
			\includegraphics[width=\textwidth]{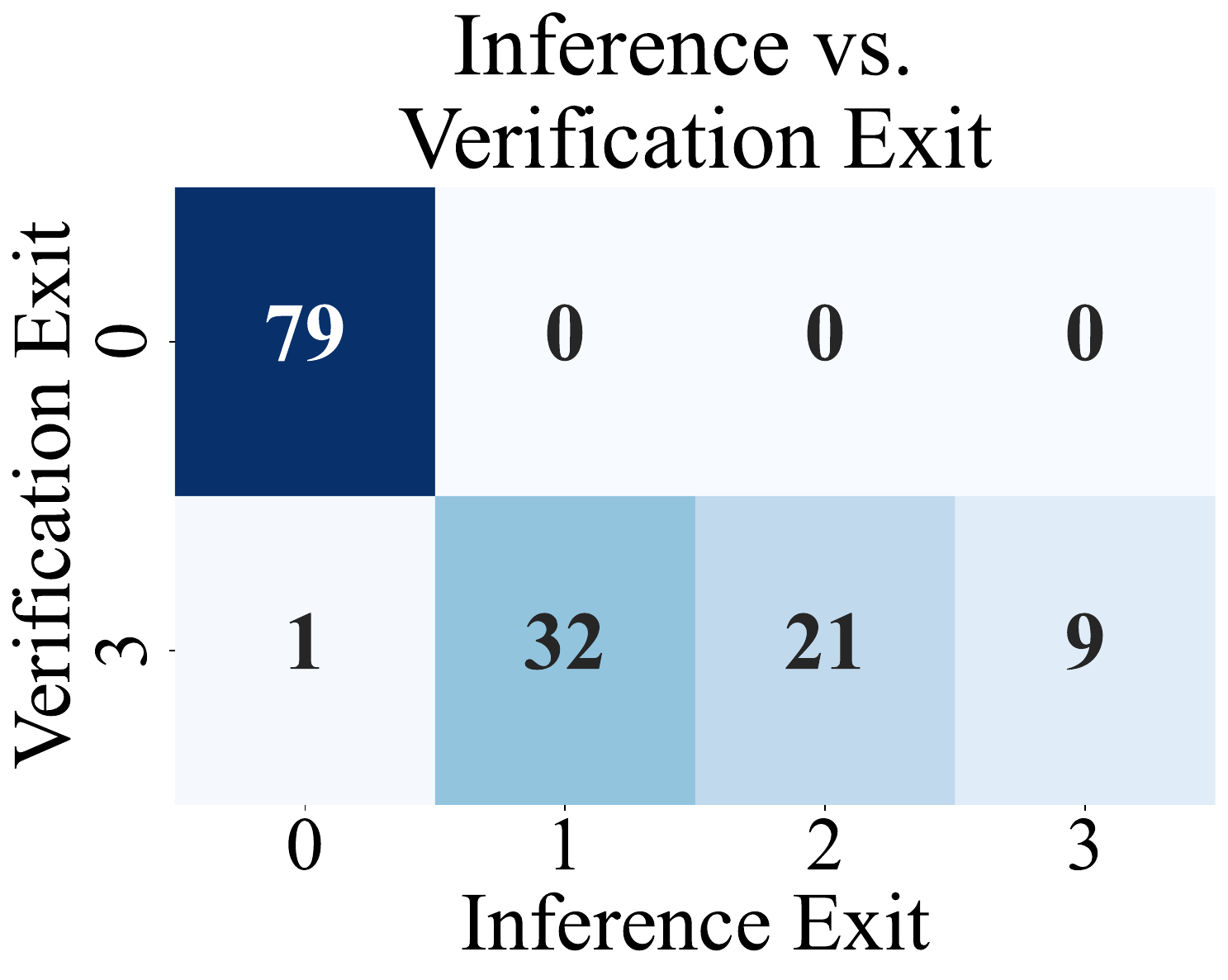}
			\caption*{\small{(e) FC-6 \safe{}}}
		\end{minipage}
		\hfill
		\begin{minipage}[b]{0.48\textwidth}
			\centering
			\includegraphics[width=\textwidth]{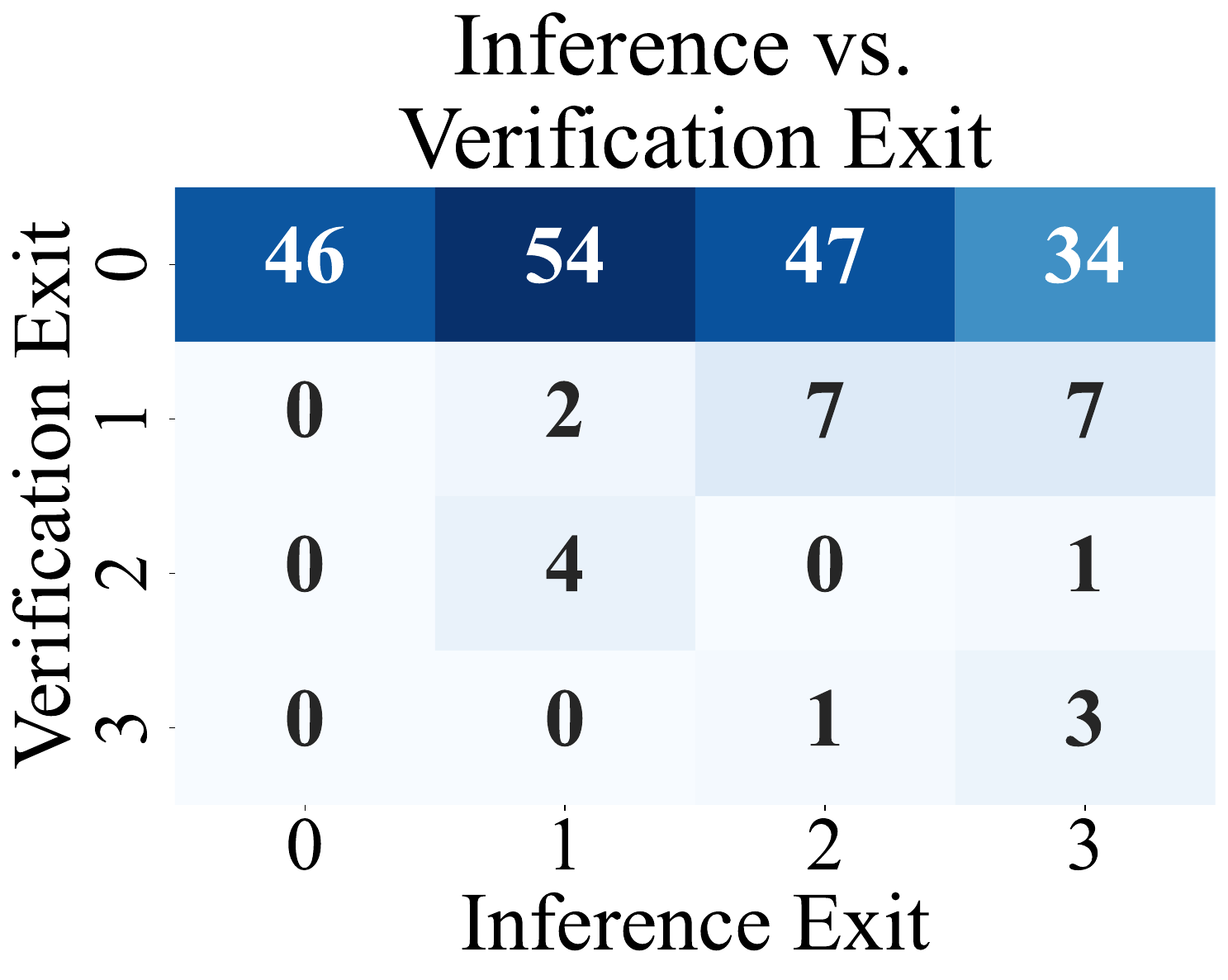}
			\caption*{\small{(f) FC-6 \unsafe{}}}
		\end{minipage}
		\caption{Heatmaps of inference vs.\ verification exit layer for FC-6 and MNIST. As with the CIFAR-10 models, a strong diagonal correlation is observed for \safe{} cases.}
		\label{fig:inf-vs-ver-exit-heatmap-mnist}
	\end{figure}

	\section{Ablation Study: Algorithm Variants}\label{app:compare-optimizations}
\begin{figure}[htbp]
		\centering
		\begin{tabular}{cc}
			\includegraphics[width=0.45\textwidth]{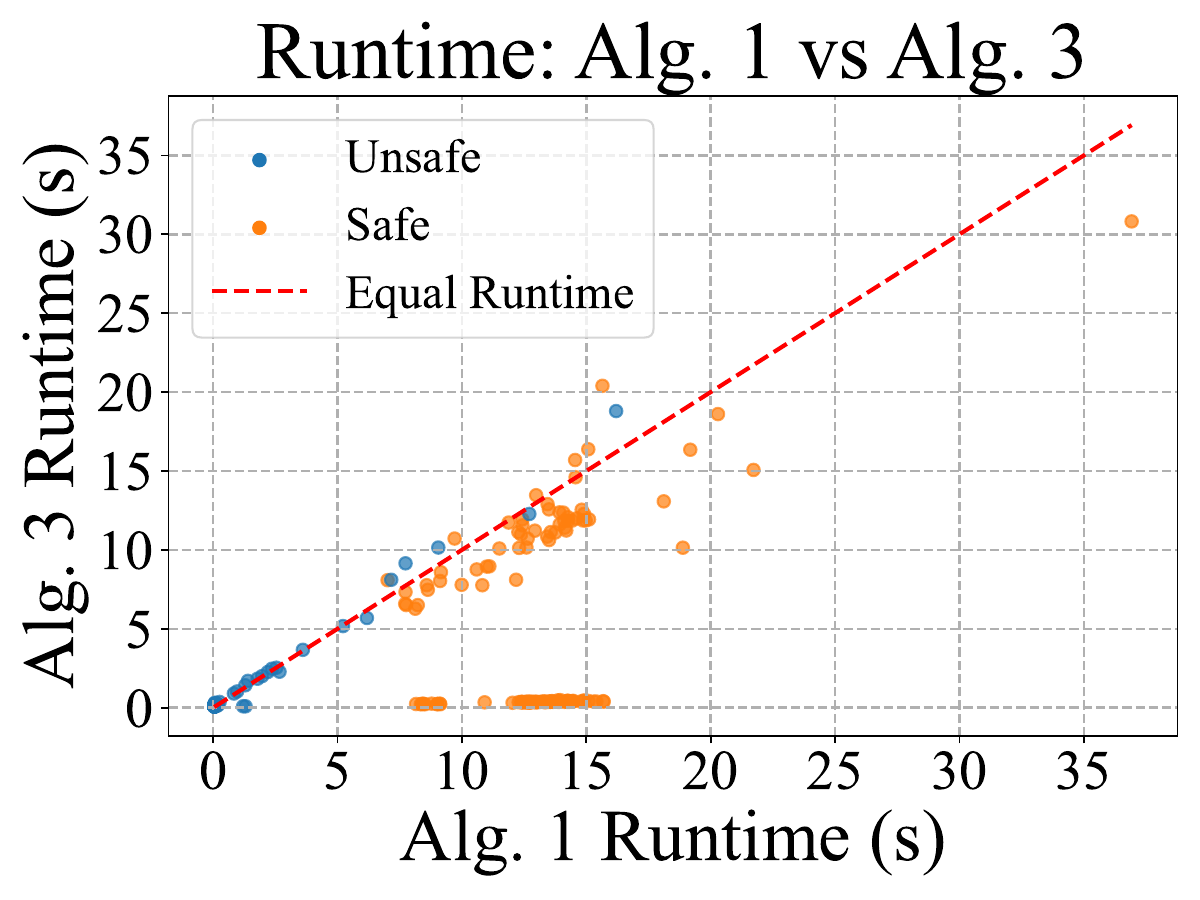} &
			\includegraphics[width=0.45\textwidth]{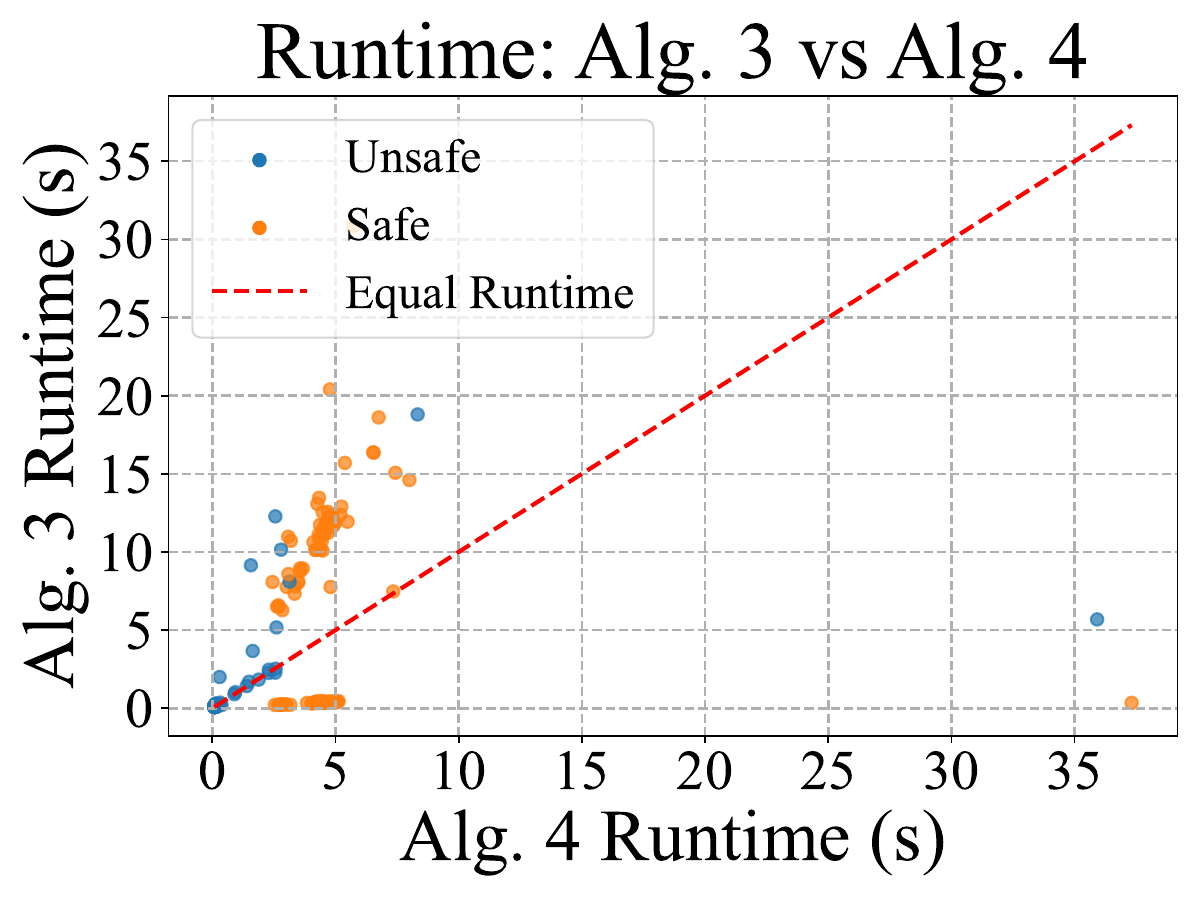} \\
			\small{(a) FC-6 MNIST: Alg.~\ref{alg:basic-early-exit-verification} vs \ref{alg:break-optimization}} & \small{(b) FC-6 MNIST: Alg.~\ref{alg:break-optimization} vs \ref{alg:continue-optimization}} \\[6pt]
			\includegraphics[width=0.45\textwidth]{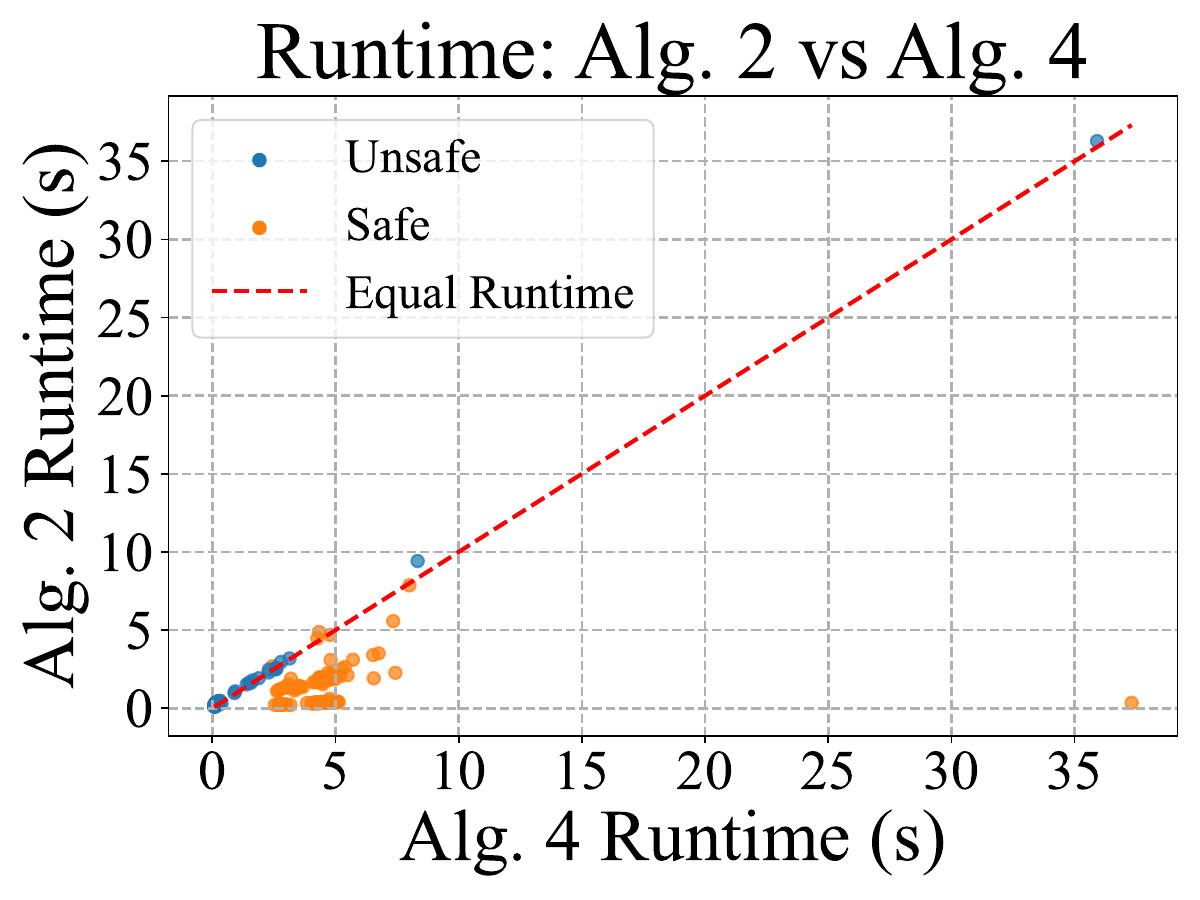} &
			\includegraphics[width=0.45\textwidth]{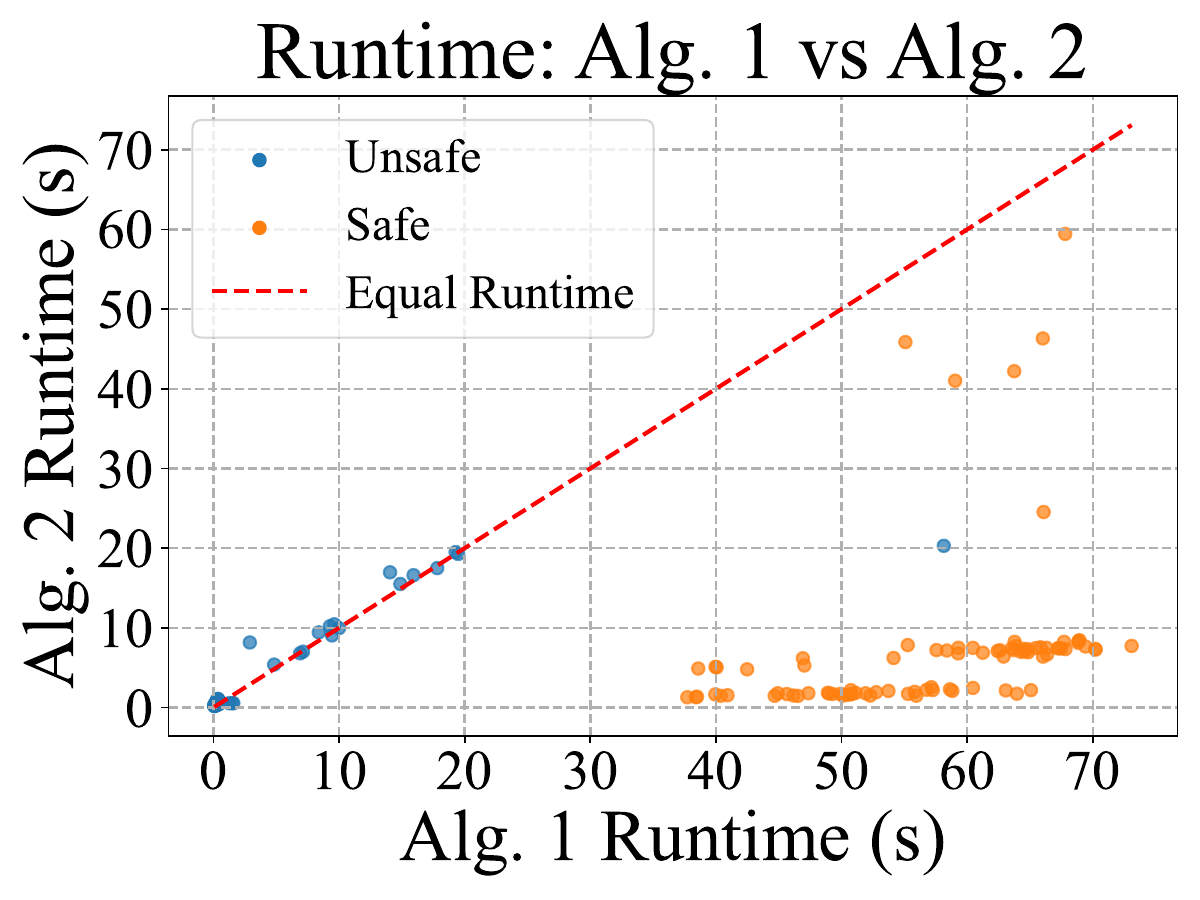} \\
			\small{(c) FC-6 MNIST: Alg.~\ref{alg:continue-optimization} vs \ref{alg:combined-break-then-continue-optimizations}} & \small{(d) LeNet-5 CIFAR-10: Alg.~\ref{alg:basic-early-exit-verification} vs \ref{alg:combined-break-then-continue-optimizations}} \\[6pt]
			\includegraphics[width=0.45\textwidth]{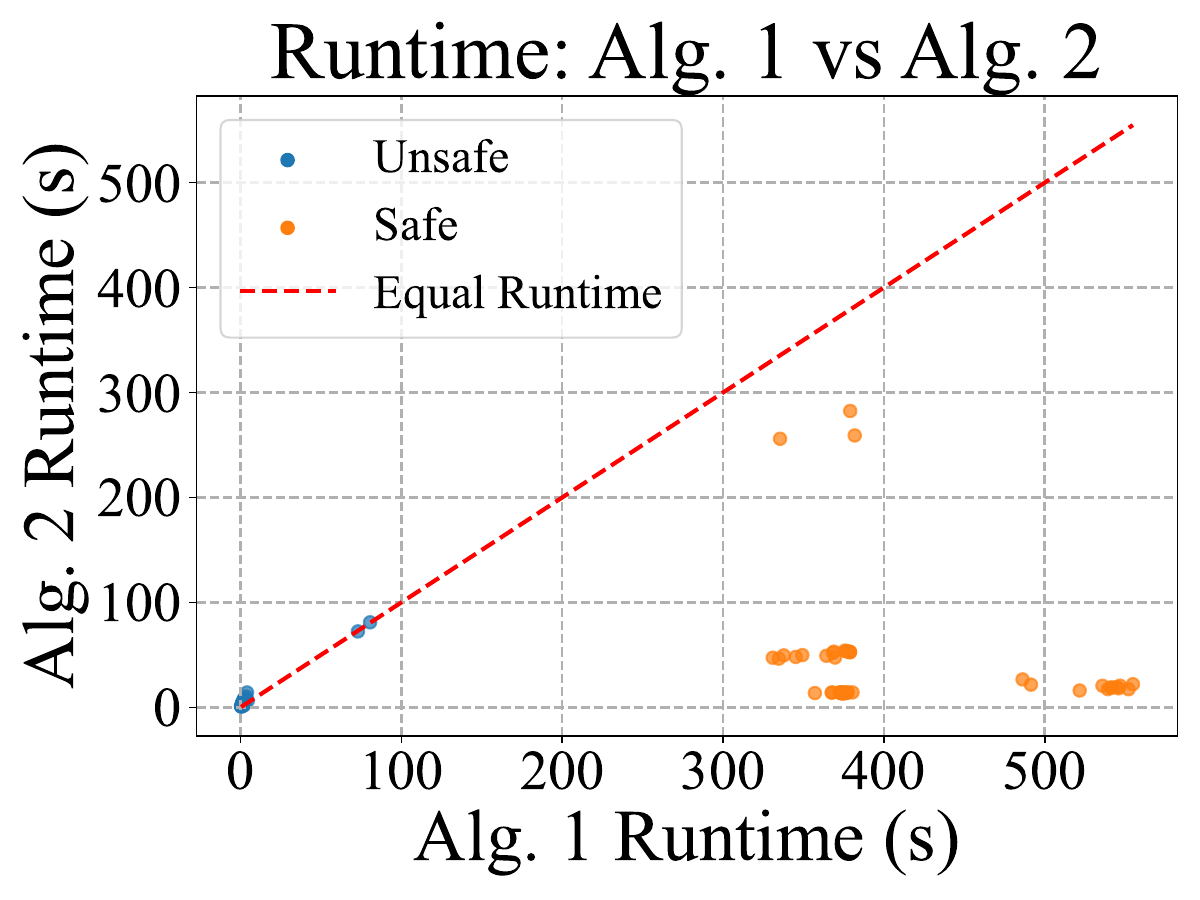} &
			\includegraphics[width=0.45\textwidth]{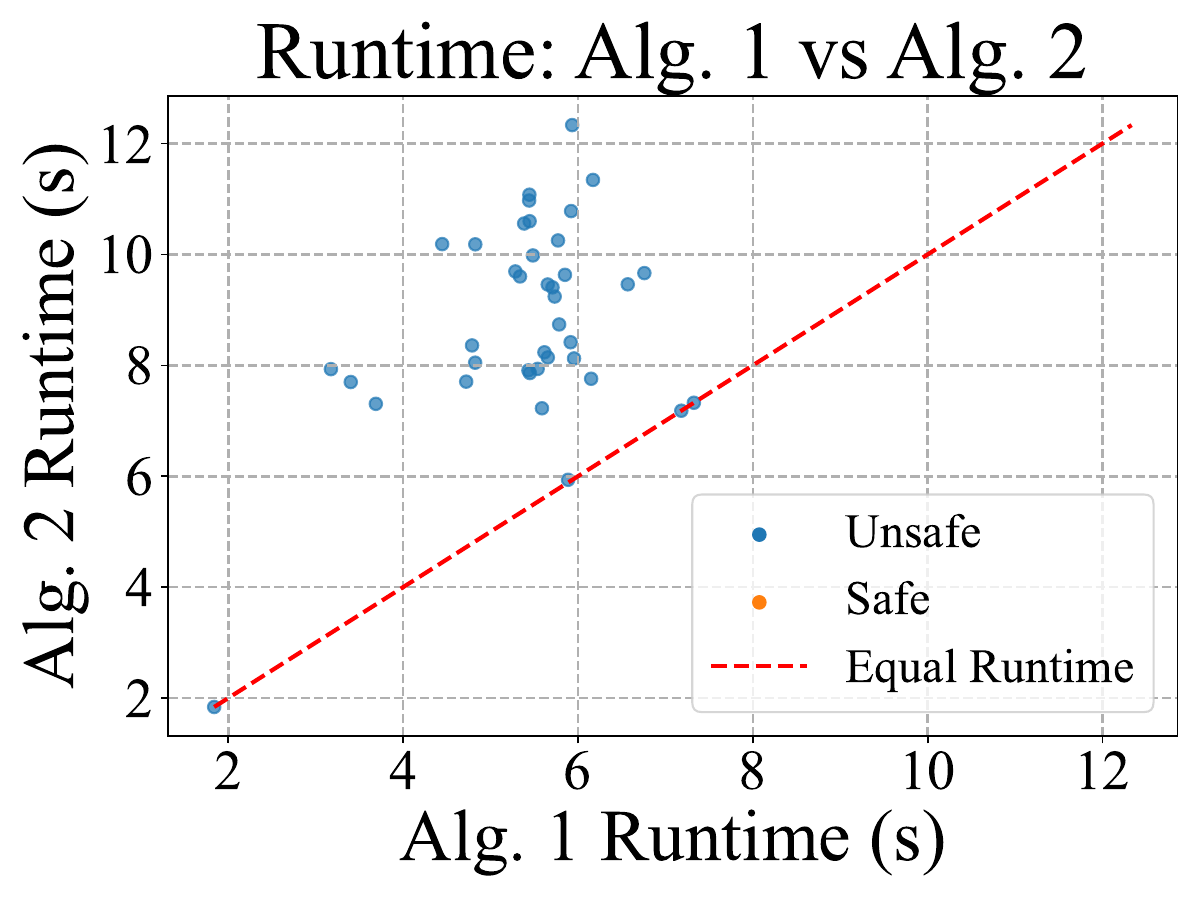} \\
			\small{(e) ResNet-18 CIFAR-10: Alg.~\ref{alg:basic-early-exit-verification} vs \ref{alg:combined-break-then-continue-optimizations}} &
			\small{(f) VGG-16 CIFAR-10: Alg.~\ref{alg:basic-early-exit-verification} vs \ref{alg:combined-break-then-continue-optimizations}} \\
		\end{tabular}
		\caption{Ablation study comparing algorithm variants across benchmarks.}
		\label{fig:compare-eev-algorithms}
	\end{figure}

	We compare \cref{alg:basic-early-exit-verification} and \cref{alg:combined-break-then-continue-optimizations} with each optimization applied individually.
	\cref{fig:compare-eev-algorithms} compares algorithm pairs; axes represent runtime (seconds); points are color-coded by result.
	
	Subfigures (a)--(c) ablate on MNIST: (a) Break-only (\cref{alg:break-optimization}) outperforms the baseline; (b) Continue-only (\cref{alg:continue-optimization}) outperforms Break-only; (c) the combined algorithm (\cref{alg:combined-break-then-continue-optimizations}) is the best. Subfigures (d)--(f) compare the baseline and combined algorithm on LeNet-5, ResNet-18, and VGG-16, showing consistent superiority in \safe{} cases and equal runtime on \unsafe{} cases.
	
	\section{Extendability to Other Properties and Activations}
	\label{app:extendability}
	While we instantiated \pave{} for local robustness, the formulation extends naturally to other property types and activation functions.
	
	\paragraph{Other Verification Properties.}
	At each exit $k$, the robustness clause can be replaced by an arbitrary predicate $\phi$ over network outputs:
	\[
	\exists x'\in B_\varepsilon(x): \phi(N_k(x')) \land \forall e<k: N_e^w(x')<T_e,
	\]
	where the right-hand term enforces the EE semantics. Examples include safety constraints ($\phi \equiv g(N_k(x')) \le 0$) and fairness predicates.
	The Break optimization generalizes by substituting the class-margin test with the corresponding sufficient condition for the new property.
	
	\paragraph{Activation Functions.}
	Our framework is activation-independent. The FPT result (\cref{the:Pee_is_FPT}) assumes piecewise-linear activations, but soundness and completeness of the general formulation hold as long as the underlying verifier supports the activation functions used.
	
	\section{Encoding Details and Choice of Underlying Verifier}\label{app:encoding-and-verifier-choice}
	
	In this appendix we elaborate on the encoding of our verification queries and explain our choice of underlying verifier.
	
	\paragraph{Structure of our queries.}
	Each verification query produced by \cref{alg:basic-early-exit-verification} or \cref{alg:combined-break-then-continue-optimizations} is a conjunction of constraints over the network outputs at \emph{multiple} exit layers. Concretely, at exit $k$, we must simultaneously assert that (i) a runner-up wins at exit $k$, and (ii) for every preceding exit $e<k$ with an EE, the winner's SoftMax score is below the threshold $T_e$. Expressing this natively requires a verifier that supports nested conjunctions ($\land$) over multiple output heads together with the SoftMax used by the EE confidence function.
	
	\paragraph{On the choice of $\alpha$-$\beta$-CROWN as the sole backend.} $\alpha$-$\beta$-CROWN is a state-of-the-art verifier that handles SoftMax robustly enough to support our queries at scale. SoftMax support is essential for our setting: without it, the threshold condition on an early exit cannot be encoded at all, since it is defined on post-SoftMax probabilities. We evaluated two alternatives: Marabou can encode SoftMax in principle, but stalls under full branch-and-bound even on the first early-exit query — a limitation confirmed by the Marabou maintainers. NeuralSAT lacks SoftMax support entirely~\citep{DuNgDw24}, making it incompatible with our encoding without tool-level modifications. As SoftMax support matures in other verifiers, our framework requires no algorithmic changes to adopt them.
	
	
	
	\paragraph{Our encoding.}
	To circumvent the lack of nested AND support in $\alpha$-$\beta$ CROWN, we decompose each multi-conjunct query into a collection of smaller single-conjunct queries and combine their results post hoc in a sound manner (a conjunction is \unsafe{} only if each conjunct is individually satisfiable under the same witness, which we enforce by the structure of the loop in \cref{alg:basic-early-exit-verification}). To avoid biasing the comparison, we apply the \emph{same} decomposition when running the baseline (no-EE) verification queries. This guarantees a fair head-to-head comparison and isolates the effect of early exits on verification time. We consulted the $\alpha$-$\beta$-CROWN maintainers, who confirmed that it is an implementation limitation -- theoretically it seems to be possible to support but currently the code only supports relatively simple specifications.
	
	\paragraph{Future simplifications.}
	Our decomposition is a workaround, not a fundamental requirement. Once a verifier supporting both SoftMax and native nested ANDs becomes available (e.g., via an extension of Marabou or $\alpha$-$\beta$ CROWN), each query in \cref{alg:basic-early-exit-verification} could be issued as a single call, removing encoding overhead and further reducing runtime. Our framework is verifier-agnostic and would directly benefit from such improvements.
	
	\section{Why Trace Stability Does Not Hold for Unsafe Queries}\label{app:trace-stability-unsafe}
	
	Recall that the trace $\tau(x)$ of an input $x$ is the set of layers $x$ traverses before an EE condition fires, and $\tau(x)$ is \emph{stable} in $B_\epsilon^x$ if $\tau(x')=\tau(x)$ for every $x'\in B_\epsilon^x$ (see \cref{subsec:3.3}). In \cref{fig:inf-vs-ver-exit-heatmap}, we observed that trace stability holds in practice for the large majority of inputs that yield \safe{} results, but breaks down on inputs that yield \unsafe{} results. Here we explain why this asymmetry is expected---and, importantly, why it does not compromise the soundness, completeness, or practical efficiency of \pave{}.
	
	\paragraph{SAFE inputs: stability is the typical regime.}
	An input $x$ for which local robustness holds in $B_\epsilon^x$ is, by definition, a ``well-separated'' input: the winner class dominates over the entire ball, so the SoftMax distribution changes only mildly under perturbation. In particular, the winner's SoftMax score tends to remain above the exit threshold $T$ at the same layer at which $x$ itself exits. Thus every $x'\in B_\epsilon^x$ produces the same trace as $x$, and trace stability holds. This is exactly what we observe empirically in \cref{fig:inf-vs-ver-exit-heatmap}(a,c): for \safe{} cases, the inference exit and verification exit layers coincide on the diagonal.
	
	\paragraph{UNSAFE inputs: stability cannot be expected to hold.}
	An \unsafe{} query is, by definition, witnessed by a perturbation $x'\in B_\epsilon^x$ whose predicted class differs from that of $x$. For the prediction to flip, the SoftMax distribution at \emph{some} exit must change substantially: either (i) the original winner's score at an early exit drops below the threshold $T$, causing $x'$ to not exit where $x$ did, or (ii) a runner-up's score rises above $T$ at an earlier exit, causing $x'$ to exit earlier. In either case, $\tau(x')\ne \tau(x)$, so $\tau(x)$ is \emph{not} stable in $B_\epsilon^x$. Trace stability and the existence of an adversarial counterexample are thus fundamentally at odds: any counterexample is a direct witness of trace instability.
	
	\paragraph{Why this is not a problem.}
	The FPT guarantee of \cref{the:Pee_is_FPT} is conditioned on trace stability and applies to \safe{} instances, which is exactly where a tight complexity bound matters (\unsafe{} instances are resolved as soon as a counterexample is found, typically in a single early-exit query---see the UNSAFE runtimes in \cref{tab:verification-stats}). Soundness and completeness of \cref{alg:basic-early-exit-verification} and \cref{alg:combined-break-then-continue-optimizations} (\cref{the:basic_verifies_Pee,the:optimized-verify_Pee}) do \emph{not} depend on trace stability at all, and thus hold on \unsafe{} inputs regardless.
	
	\paragraph{Empirical corroboration.}
	The heatmaps in \cref{fig:inf-vs-ver-exit-heatmap}(b,d) and \cref{fig:inf-vs-ver-exit-heatmap-mnist} show that \unsafe{} queries are concentrated in the first verification exit: counterexamples, when they exist, are typically found early regardless of where the original input $x$ would have exited. This matches the theoretical picture above: the counterexample takes a different path through the network than $x$, and our algorithm discovers it as soon as that path is explored.
	
	\section{Disclosure: Use of Large Language Models (LLMs)}\label{app:llm_use_disclosure}
	The authors were solely responsible for developing the research questions, designing the methodology, performing the analysis, and interpreting the findings. A large language model (LLM) was employed only to assist with improving the clarity and style of the writing, without influencing any substantive aspects of the research.

	

\end{document}